\documentclass[11pt]{article}

\usepackage[preprint]{acl}

\usepackage{times}
\usepackage{latexsym}
\usepackage[T1]{fontenc}
\usepackage[utf8]{inputenc}
\usepackage{inconsolata}
\usepackage{graphicx}

\usepackage{amsmath}
\usepackage{amssymb}
\usepackage{booktabs}
\usepackage{multirow}

\newif\ifincludeappendix
\includeappendixtrue

\title{Dual-Side Uncertainty Estimation:\\Calibrating Data and Model Uncertainty into a Shared Probability Space}%

\author{
  Jiaxuan Gao\textsuperscript{1},
  Yang Hu\textsuperscript{1},
  Feihong Lu\textsuperscript{2},
  Yudi Huang\textsuperscript{1},
  \textbf{Qingyun Sun\textsuperscript{1\thanks{Corresponding author}}} \and
  \textbf{Jianxin Li\textsuperscript{1}} \\[2pt]
  \textsuperscript{1}Beihang University, School of Computer Science and Engineering, Beijing, China \\
  \textsuperscript{2}Zhengzhou University, School of Cyber Science and Engineering, Zhengzhou, China
}

\hypersetup{
  pdftitle={Dual-Side Uncertainty Estimation: Calibrating Data and Model Uncertainty into a Shared Probability Space},
  pdfauthor={Jiaxuan Gao, Yang Hu, Feihong Lu, Yudi Huang, Qingyun Sun, Jianxin Li},
  pdfsubject={Multimodal intention analysis and uncertainty estimation},
  pdfkeywords={uncertainty estimation, calibration, multimodal learning, language models}
}

\begin{document}

\maketitle

\begin{abstract}

Multimodal user intention analysis infers latent goals from heterogeneous behavioral observations but faces cross-modal input mismatch and output semantic errors. Existing uncertainty estimators focus on input reliability or output confidence and often reduce token uncertainty to simple statistics, overlooking structured dependencies. We propose DUE (Dual-Side Uncertainty Estimation), which jointly estimates input-side aleatoric and sentence-level epistemic uncertainty. From a single generation, Dirichlet perturbations of cross-layer representations produce an ordered token uncertainty sequence. A sequence-aware Normal-Inverse-Gamma estimator uses this sequence to quantify epistemic uncertainty and confidence in its estimate. Evidential multimodal alignment captures aleatoric uncertainty from noisy, incomplete, or inconsistent inputs. DUE calibrates both heterogeneous scores into a shared error-probability space and sequentially fuses them to estimate overall generation risk. Experiments demonstrate strong performance in input-mismatch detection, output semantic-error detection, and overall reliability estimation without repeated generation.
\end{abstract}

\begin{figure}[!t]
    \centering
    \includegraphics[width=0.95\columnwidth]{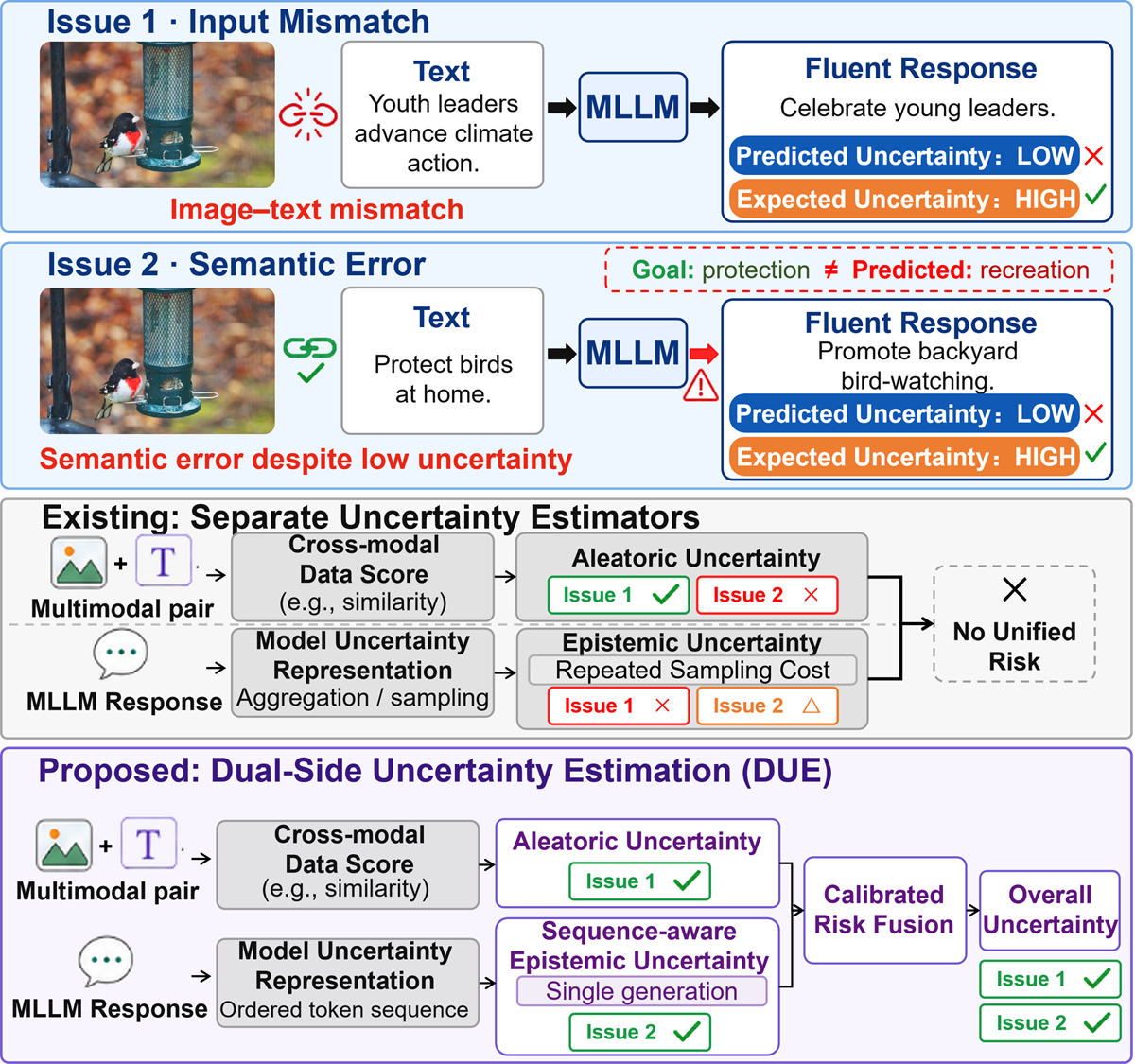}
    \caption{Motivation of our framework. Input mismatch (issue 1) and semantic generation errors (issue 2) represent complementary sources of unreliability that may be overlooked by input or output only estimators. DUE calibrates both risks into comparable error probabilities and sequentially fuses them into a unified uncertainty estimate from a single response. }
    \label{fig:motivation}
\end{figure}

\section{Introduction}
Multimodal user intention analysis infers latent goals from behavioral observations, supporting social understanding~\cite{zadeh2019socialiq,lu2026mina}, personalized recommendation~\cite{zhou2018din,fang2026mimar}, and safety-critical decision making~\cite{girase2021loki,qu2025efficientpie}.
Recent methods, such as MIND~\cite{xu2024mind} and MINA~\cite{lu2026mina}, leverage complementary multimodal cues, yet these cues may be incomplete or inconsistent.
As shown in Figure~\ref{fig:motivation}, unreliable intention inference may arise from either deficient cross-modal evidence or erroneous semantic reasoning despite reliable inputs.

Uncertainty estimation provides a principled way to characterize these
two sources. Aleatoric uncertainty reflects ambiguity or inconsistency in
multimodal evidence, whereas epistemic uncertainty reflects limitations
in model knowledge or inference. Existing epistemic uncertainty
estimators face a trade-off between efficiency and semantic reliability.
Token-level methods derive uncertainty from predictive distributions or
internal representations, but simple aggregation ignores token order,
sequential dependencies, and semantically critical positions
~\cite{malinin2021uncertainty,vashurin2025benchmarking}.
FLUE~\cite{gao2025flue} models transitions in hidden-state-derived token
uncertainty but produces a deterministic sentence score without
quantifying estimation confidence. Semantic Entropy
~\cite{kuhn2023semantic} and SelfCheckGPT
~\cite{manakul2023selfcheckgpt} capture semantic variation but require
multiple generations. Moreover, token-distribution uncertainty does not
necessarily indicate semantic error. Reliable sentence-level estimation
therefore requires structured sequence modeling, confidence
quantification, and explicit semantic alignment.

Aleatoric uncertainty methods assess ambiguous, noisy, or inconsistent multimodal evidence through probabilistic representations or reliability signals~\cite{chun2024improved,fang2023uatvr,li2023prototype,duan2025fuzzy}, but primarily target matching or retrieval rather than epistemic risk in open-ended generation. These distinct risks~\cite{kendall2017uncertainties,bickfordsmith2025rethinking} are typically estimated independently, with raw scores differing in scale, distribution, and event semantics. Direct aggregation may therefore obscure interpretation or suppress high risk from either source. Their sequential input-to-generation failure process motivates event-specific calibration~\cite{guo2017calibration} and corresponding fusion.

These estimation and integration gaps motivate \textbf{DUE}, a
\textbf{D}ual-side \textbf{U}ncertainty \textbf{E}stimation framework for
reliable multimodal intention analysis. DUE decomposes generation
unreliability into input-side aleatoric uncertainty and sentence-level
epistemic uncertainty. The aleatoric branch evaluates cross-modal
consistency and evidential sufficiency, while the epistemic branch
constructs an ordered token uncertainty sequence from a single generated
response for sequence-aware sentence-level estimation. The two scores are calibrated into comparable event-specific error probabilities and sequentially fused to yield an interpretable overall reliability estimate.
Our main contributions are as follows:
\begin{itemize}
    \item We propose a single-generation epistemic uncertainty estimator that models structured dependencies among token-level uncertainties for sentence-level estimation.
    \item We develop a dual-side framework that calibrates input-side aleatoric and sentence-level epistemic uncertainty into a shared error-probability space for overall generation-risk estimation.
    \item Experiments demonstrate state-of-the-art discrimination of input mismatch, semantic generation errors, and overall unreliability using only a single generation.

\end{itemize}

\section{Related Work}

\paragraph{Multimodal Intention Understanding.}
User intention is often implicit and emerges from cross-modal
interactions~\cite{kruk2019integrating}. Recent work studies shopping and
social-media intentions with MIND~\cite{xu2024mind} and
MINA~\cite{lu2026mina}, and safety-critical traffic intentions with
LOKI~\cite{girase2021loki} and CIPF~\cite{ham2023cipf}. General-purpose
models such as LLaVA and MiniGPT-4 support broader multimodal reasoning
~\cite{liu2023visual,zhu2024minigpt4}. Yet multimodal models may over-rely
on text~\cite{mullick2025text}, and few studies separate input
unreliability from semantic inference errors.

\paragraph{Uncertainty Estimation for Autoregressive Generation.}
Autoregressive uncertainty is estimated from token probabilities,
ensembles, or sequence decompositions~\cite{malinin2021uncertainty}, with
recent benchmarks comparing these approaches
~\cite{vashurin2025benchmarking}. Verbalized confidence may not reflect
semantic correctness~\cite{kadavath2022language}, while Semantic Entropy
~\cite{kuhn2023semantic} and SelfCheckGPT
~\cite{manakul2023selfcheckgpt} require repeated decoding. Efficient
alternatives use hidden-state geometry~\cite{chen2024inside}, uncertainty
probes~\cite{kossen2024sep}, token sequences~\cite{gao2025flue},
cross-layer dynamics~\cite{zhang2025icr}, or reusable heads
~\cite{shelmanov2025head}. Most yield point estimates without
sentence-level confidence. DUE integrates ordered token-level uncertainty sequences with sequence-aware NIG learning and semantic supervision.

\begin{figure*}[t]
    \centering
    \includegraphics[width=0.9\textwidth]{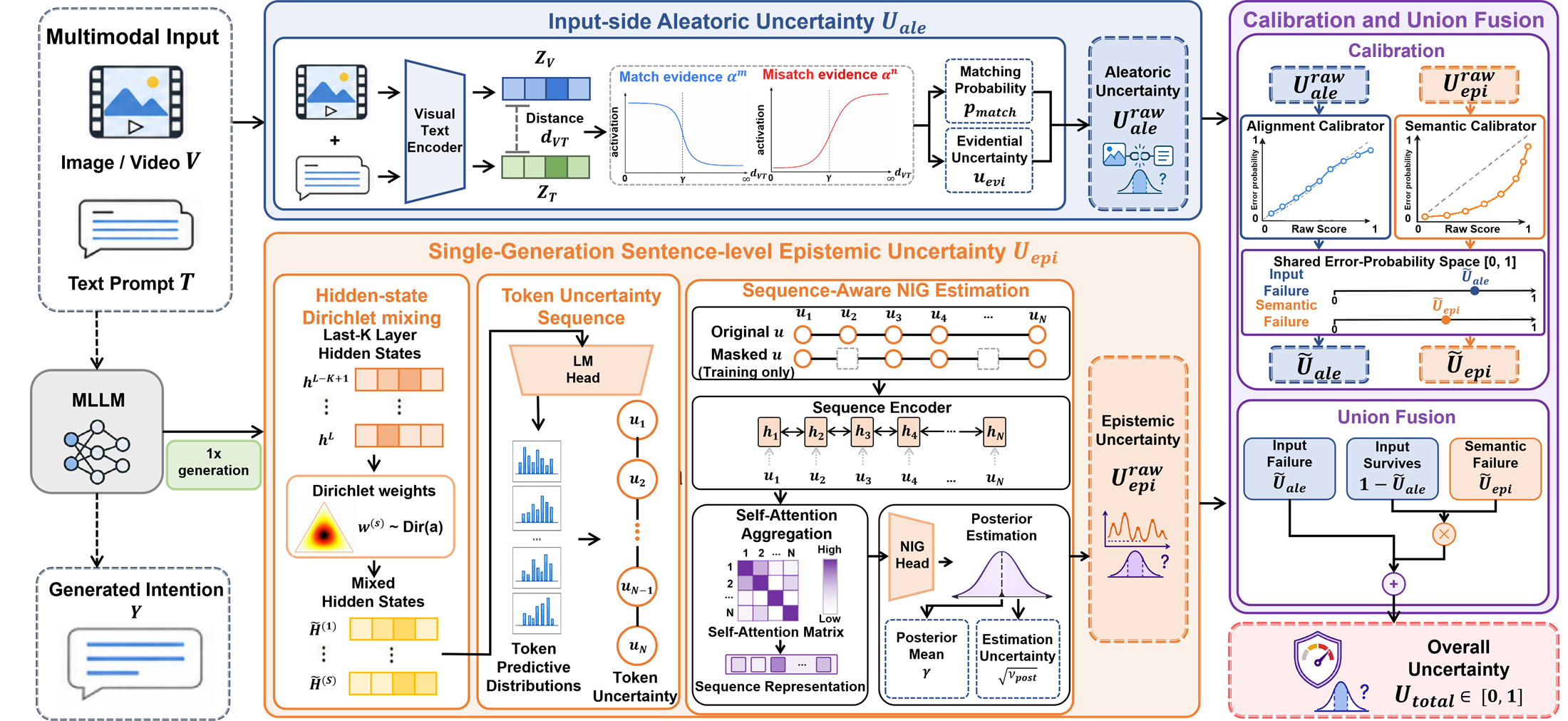}
    \caption{Overview of DUE. It decomposes generation unreliability into input-side aleatoric risk from ambiguous or inconsistent multimodal evidence and sentence-level epistemic risk from structured token uncertainty in a single response.  Event-specific calibration converts both into comparable error probabilities, which are fused by sequential Noisy-OR along the input-to-generation failure process.}
    \label{fig:framework}
\end{figure*}

\paragraph{Aleatoric and Epistemic Uncertainty in Multimodal Models.}
Cross-modal uncertainty extends deterministic CLIP alignment
~\cite{radford2021learning} through probabilistic embeddings
~\cite{chun2021probabilistic,chun2024improved}, uncertainty-aware
pre-training, retrieval, and prototype learning
~\cite{ji2023map,fang2023uatvr,li2023prototype}, and fuzzy retrieval
~\cite{duan2025fuzzy}. Related work also estimates hallucination risk
~\cite{zhang2024vluncertainty} or uses visual evidence for
self-evaluation~\cite{park2026vauq}. Evidential learning models evidence
and ignorance~\cite{sensoy2018evidential} through NIG regression
~\cite{amini2020deep}, multimodal fusion~\cite{ma2021trustworthy}, and
confidence under missing modalities~\cite{ma2023calibrating}, while
post-hoc calibration maps scores to probabilities
~\cite{guo2017calibration}. DUE jointly calibrates and sequentially fuses
input-side aleatoric and sentence-level epistemic risks.

\section{Problem Formulation}
\label{sec:problem}

Given visual input $V$ and text $T$, an MLLM generates an intention $Y$,
with reference $Y^*$ available during training and evaluation. We
consider two sequential failure events. Input failure
$\mathcal E_{\mathrm{align}}$ occurs when the multimodal evidence is
inconsistent
, whereas semantic failure
$\mathcal E_{\mathrm{sem}}$ occurs when $Y$ is semantically incorrect
despite reliable evidence. DUE associates these events with aleatoric
risk $U_{\mathrm{ale}}$ and epistemic risk $U_{\mathrm{epi}}$,
respectively.

To assign each failure to a unique stage, the overall failure event is
defined as
\begin{equation}
\small
\mathcal E_{\mathrm{total}}
=
\mathcal E_{\mathrm{align}}
\cup
(\neg\mathcal E_{\mathrm{align}}
\cap\mathcal E_{\mathrm{sem}}).
\label{eq:total_event}
\end{equation}
The goal is to estimate the overall failure probability
$U_{\mathrm{total}}\approx P(\mathcal E_{\mathrm{total}}\mid V,T,Y)$
by calibrating and sequentially fusing both scores.

\section{Method}
\label{sec:method}

DUE separates aleatoric risk due to unreliable multimodal evidence from
epistemic risk during semantic inference, enabling source-specific
assessment. Figure~\ref{fig:framework} shows its aleatoric and
sentence-level epistemic estimators, event-specific calibration, and
sequential fusion.

Drawing on evidential learning~\cite{sensoy2018evidential}, the aleatoric
estimator evaluates evidence consistency and sufficiency under noisy,
incomplete, or inconsistent inputs. The epistemic estimator constructs an
ordered token uncertainty sequence through Dirichlet perturbations of
cross-layer representations from a single generation. Sequence-aware NIG
modeling~\cite{amini2020deep} captures token dependencies and jointly
estimates sentence-level epistemic uncertainty and posterior estimation
uncertainty. Masked prediction learns sequential structure, while
classification and ranking supervision align the score with semantic-error
occurrence and severity. Separate calibrators then map both scores to
event-specific error probabilities, which are sequentially fused to
estimate overall reliability.
\subsection{Input-Side Aleatoric Uncertainty}
\label{sec:input_uncertainty}

The aleatoric estimator determines whether $V$ and $T$ provide
consistent and sufficient evidence. A vision--language encoder first
produces normalized representations $z_V$ and $z_T$, from which we
compute the projected cosine similarity
$c_{VT}=z_V^\top z_T$ and Euclidean distance
$d_{VT}=\lVert z_V-z_T\rVert_2$. A shared pair encoder then forms
\begin{equation}
\small
\mathbf h_{VT}=\rho([c_{VT},d_{VT}]),
\label{eq:decoupled_evidence}
\end{equation}
where $\rho$ is an MLP. From $\mathbf h_{VT}$, two decoupled heads
estimate the direction and strength of the available evidence:
\begin{align}
\small
\boldsymbol\pi_{VT}
&=\operatorname{Softmax}
(\mathbf W_{\pi}\mathbf h_{VT}+\mathbf b_{\pi}),
\label{eq:evidence_direction}\\
s_{VT}
&=\kappa\operatorname{Softplus}
(\mathbf w_s^\top\mathbf h_{VT}+b_s).
\label{eq:evidence_strength}
\end{align}
Here, $\boldsymbol\pi_{VT}=(\pi^m_{VT},\pi^n_{VT})$ specifies whether
the evidence supports matching or mismatching, while $s_{VT}$ measures
its total strength and $\kappa$ controls its scale.

For class $c\in\{m,n\}$, the directional evidence and corresponding
Dirichlet parameter are
\begin{align}\label{eq:class_evidence}
\small
e^c_{VT}=s_{VT}\times\pi^c_{VT},
\alpha^c_{VT}=e^c_{VT}+1.
\end{align}
Let $S_{VT}=\alpha^m_{VT}+\alpha^n_{VT}$. We use
$p_{\mathrm{match}}=\pi^m_{VT}$ for the matching direction and
$u_{\mathrm{evi}}=2/S_{VT}$ for residual ignorance. Their combination
defines the evidential aleatoric risk:
\begin{equation}
\small
U_{\mathrm{ale}}^{\mathrm{evi}}
=1-p_{\mathrm{match}}(1-u_{\mathrm{evi}}).
\label{eq:uev_align}
\end{equation}
This score rises with mismatch or insufficient evidence, preventing weak
support from appearing confidently matched.

The learned evidence is complemented with raw encoder cues. Let
$c_{\mathrm{raw}}$ denote visual--text cosine similarity and
$\mathbf c=(c_1,\ldots,c_F)$ the frame--text similarities. We collect
the 13-dimensional feature vector
\begin{equation}
\begin{aligned}
\mathbf f_{VT} = \big[
&U_{\mathrm{ale}}^{\mathrm{evi}},
\boldsymbol{\pi}_{VT},u_{\mathrm{evi}},
s_{VT},c_{VT},d_{VT},\\
&c_{\mathrm{raw}},1-c_{\mathrm{raw}},
\boldsymbol{\phi}(\mathbf c)
\big],
\end{aligned}
\label{eq:aleatoric_features}
\end{equation}
where $\mathbf c=[c_1,\ldots,c_K]$ denotes the confidence scores of $K$ candidate predictions.

A pairwise risk head converts these complementary cues into the raw
aleatoric score:
\begin{equation}
U_{\mathrm{ale}}^{\mathrm{raw}}=g_\phi(\mathbf f_{VT}).
\label{eq:raw_align}
\end{equation}

\subsection{Sentence-Level Epistemic Uncertainty}
\label{sec:output_uncertainty}

The epistemic estimator preserves the order of token uncertainties from
a single generation. For token $t$, we collect the last $K$ hidden
states and independently sample
$\mathbf w_t^{(s)}\sim\operatorname{Dir}(\mathbf a)$. The sampled
representation is
\begin{equation}
\widetilde h_t^{(s)}
=\sum_{k=1}^{K}w_{tk}^{(s)}h_t^{L-K+k}.
\label{eq:hidden_mix}
\end{equation}
It is mapped through the original final normalization and language-model
head to obtain
\begin{equation}
p_t^{(s)}
=\operatorname{Softmax}\!
\bigl(
\operatorname{LMHead}\!
\bigl(
\operatorname{FinalNorm}\!(\widetilde h_t^{(s)})
\bigr)
\bigr).
\label{eq:mixed_prediction}
\end{equation}
Because the Dirichlet weights are nonnegative and sum to one, each
mixture lies in the convex hull of the selected high-layer states. Under
the assumptions in the  supplementary material, Propositions A.1--A.3 establish
retained next-token information, bounded perturbations, and convergence
to the induced predictive entropy. The mixing therefore probes local
representation ambiguity without additional decoding.

The sampled distributions are first averaged:
\begin{equation}
\bar p_t=\frac{1}{S}\sum_{s=1}^{S}p_t^{(s)}.
\label{eq:average_prediction}
\end{equation}
Their predictive entropy defines token uncertainty:
\begin{equation}
u_t=-\sum_{v\in\mathcal V}\bar p_t(v)\log\bar p_t(v).
\label{eq:token_entropy}
\end{equation}
For sample $i$, these values form the ordered sequence
$\mathbf u_i=(u_{i1},\ldots,u_{iN_i})$, normalized using statistics
estimated from the training split.

Simple mean or maximum aggregation discards token positions and
dependencies. We instead divide valid positions during training into
observed set $\mathcal O_i$ and masked set $\mathcal M_i$. For normalized
position $r_{it}=t/N_i$, the token input is
\begin{equation}
\mathbf x_{it}=[m_{it}^{O}u_{it},\,m_{it}^{O},\,r_{it}],
\label{eq:masked_token_input}
\end{equation}
where $m_{it}^{O}$ marks observed positions. A bidirectional sequence
encoder, attention, and global statistics produce
\begin{equation}
\mathbf r_i=
\big[
\operatorname{Attn}(\operatorname{BiLSTM}(\mathbf x_i)),\,
\varphi(\mathbf u_i^{\mathcal O_i})
\big],
\label{eq:sequence_representation}
\end{equation}
where $\varphi(\cdot)$ summarizes global sequence statistics.

Following deep evidential regression~\cite{amini2020deep},
$\mathbf r_i$ parameterizes a sequence-conditioned
Normal-Inverse-Gamma (NIG) prior:
\begin{equation}
(\gamma_i^0,\nu_i^0,\alpha_i^0,\beta_i^0)=f_\theta(\mathbf r_i),
\label{eq:nig_params}
\end{equation}
with $\nu_i^0>0$, $\alpha_i^0>1$, and $\beta_i^0>0$. The hierarchy
separates latent sentence uncertainty from residual token variation:
\begin{align}
\sigma_i^2\mid\mathbf u_i^{\mathcal O_i}
&\sim\operatorname{InvGamma}(\alpha_i^0,\beta_i^0),
\label{eq:nig_residual_prior}\\
\mu_i\mid\sigma_i^2,\mathbf u_i^{\mathcal O_i}
&\sim\mathcal N
\left(\gamma_i^0,\frac{\sigma_i^2}{\nu_i^0}\right),
\label{eq:nig_mean_prior}\\
u_{it}\mid\mu_i,\sigma_i^2
&\sim\mathcal N(\mu_i,\sigma_i^2).
\label{eq:latent_sentence_uncertainty}
\end{align}

For the final estimate, the prior is updated using all valid tokens.
With validity mask $m_{it}$, the sufficient statistics are
\begin{align}
n_i&=\sum_t m_{it},
\qquad
\bar u_i=\frac{\sum_t m_{it}u_{it}}{n_i},
\label{eq:token_statistics}\\
Q_i&=\sum_t m_{it}(u_{it}-\bar u_i)^2.
\label{eq:token_scatter}
\end{align}
Conjugacy gives the posterior parameters
\begin{align}
\nu_i^+&=\nu_i^0+n_i,
\qquad
\alpha_i^+=\alpha_i^0+\frac{n_i}{2},
\label{eq:nig_posterior_update}\\
\gamma_i^+
&=\frac{\nu_i^0\gamma_i^0+n_i\bar u_i}{\nu_i^+},
\label{eq:nig_posterior_mean}\\
\beta_i^+
&=\beta_i^0+\frac{Q_i}{2}
+\frac{\nu_i^0n_i(\bar u_i-\gamma_i^0)^2}{2\nu_i^+}.
\label{eq:nig_posterior_scale}
\end{align}
The posterior mean $\gamma_i^+$ estimates sentence-level epistemic
uncertainty, while its estimation variance is
\begin{equation}
v_{\mathrm{post},i}
=\frac{\beta_i^+}{\nu_i^+(\alpha_i^+-1)}.
\label{eq:posterior_variance}
\end{equation}
We combine the estimate and its confidence conservatively:
\begin{equation}
U_{\mathrm{epi},i}^{\mathrm{raw}}
=\operatorname{clip}_{[0,1]}
\left(\gamma_i^++\eta\sqrt{v_{\mathrm{post},i}}\right),
\label{eq:sentence_uncertainty_score}
\end{equation}
where $\eta$ controls the contribution of posterior uncertainty. Masked
positions define the NIG likelihood during training, whereas inference
uses the complete valid sequence.

Semantic supervision aligns uncertainty with intention-level errors.
Given $Y_i$ and reference $Y_i^*$, BERTScore
~\cite{zhang2020bertscore} defines severity
$q_i=1-\operatorname{BERTScore}(Y_i,Y_i^*)$ and binary label
$e_i=\mathbb I[\operatorname{BERTScore}(Y_i,Y_i^*)<\tau]$, where $\tau$
is selected on validation data.

The NIG posterior also yields residual token uncertainty:
\begin{equation}
v_{\mathrm{token},i}
=\mathbb E[\sigma_i^2\mid\mathbf u_i]
=\frac{\beta_i^+}{\alpha_i^+-1}.
\label{eq:token_variance}
\end{equation}
Unlike empirical variation across positions, this quantity represents
residual variation under the sentence-level posterior.

An auxiliary semantic-error head combines the uncertainty estimates with
global sequence statistics. Its input is
\begin{equation}
\mathbf s_i=[
U_{\mathrm{epi},i}^{\mathrm{raw}},
v_{\mathrm{post},i},
v_{\mathrm{token},i},
\varphi(\mathbf u_i)].
\label{eq:semantic_features}
\end{equation}
The corresponding error probability is
\begin{equation}
\widehat e_i=\sigma(c_\omega(\mathbf s_i)).
\label{eq:semantic_head}
\end{equation}

\subsection{Calibration and Union Fusion}
\label{sec:calibration_fusion}

The two raw scores describe different events and need not share the same
scale or distribution. After orienting them so that larger values indicate
greater risk, separate monotonic calibrators
~\cite{guo2017calibration} fitted on held-out data produce
\begin{align}
\small
\widetilde U_{\mathrm{ale}}=f_{\mathrm{ale}}(U_{\mathrm{ale}}^{\mathrm{raw}}),
\widetilde U_{\mathrm{epi}}=f_{\mathrm{epi}}(U_{\mathrm{epi}}^{\mathrm{raw}}).
\label{eq:epistemic_calibration}
\end{align}
The aleatoric calibrator uses input-mismatch labels. The epistemic
calibrator is fitted only on aligned samples, so it estimates semantic
failure conditional on reliable evidence.

Following Equation~\ref{eq:total_event}, the calibrated probabilities are
fused by Sequential Noisy-OR:
\begin{equation}
U_{\mathrm{total}}
=\widetilde U_{\mathrm{ale}}
+(1-\widetilde U_{\mathrm{ale}})
\widetilde U_{\mathrm{epi}}.
\label{eq:union_uncertainty}
\end{equation}
The first term captures input failure, while the second captures semantic failure after input survival. Under the sequential definition, events are non-overlapping, keeping the score in $0,1$ without assuming unconditional independence. At inference, scores from the input and one response are calibrated and fused without regeneration.

\subsection{Training Objectives}
\label{sec:total_loss}

The aleatoric and epistemic estimators are optimized separately before
held-out probability calibration.

\paragraph{Aleatoric objective.}
The aleatoric estimator uses two-stage training. First, the
visual--textual representations and decoupled evidence heads are learned
with the symmetric contrastive objective
\begin{equation}
\small
\mathcal L_{\mathrm{NCE}}
=\frac{1}{2}
\left[
\operatorname{CE}
\left(
\frac{\mathbf Z_V\mathbf Z_T^\top}{\tau_c},\mathbf I
\right)
+
\operatorname{CE}
\left(
\frac{\mathbf Z_T\mathbf Z_V^\top}{\tau_c},\mathbf I
\right)
\right].
\label{eq:input_nce_loss}
\end{equation}
Evidence direction is supervised on matched and in-batch mismatched
pairs:
\begin{equation}
\small
\mathcal L_{\mathrm{cls}}
=-\sum_{(i,j)\in\mathcal B}w_{ij}
\sum_{c\in\{m,n\}}y_{ijc}\log\pi^c_{ij},
\label{eq:evidence_classification_loss}
\end{equation}
where $w_{ij}$ emphasizes hard negatives and suppresses potential false
negatives.

Following evidential regularization~\cite{sensoy2018evidential}, evidence
for the target class is removed before regularizing unsupported evidence:
\begin{equation}
\small
\widetilde{\boldsymbol\alpha}_{ij}
=\mathbf y_{ij}
+(\mathbf 1-\mathbf y_{ij})\odot\boldsymbol\alpha_{ij}.
\label{eq:label_aware_alpha}
\end{equation}
The resulting label-aware regularizer is
\begin{equation}
\small
\mathcal L_{\mathrm{KL}}
=\sum_{(i,j)\in\mathcal B}
D_{\mathrm{KL}}
\left[
\operatorname{Dir}
(\cdot\mid\widetilde{\boldsymbol\alpha}_{ij})
\middle\|
\operatorname{Dir}(\cdot\mid\mathbf 1)
\right].
\label{eq:evi_kl_loss}
\end{equation}
The first-stage objective is
\begin{equation}
\mathcal L_{\mathrm{backbone}}
=\mathcal L_{\mathrm{NCE}}
+\lambda_{\mathrm{evi}}\mathcal L_{\mathrm{cls}}
+\lambda_{\mathrm{KL}}(t)\mathcal L_{\mathrm{KL}},
\label{eq:input_backbone_loss}
\end{equation}
where $\lambda_{\mathrm{KL}}(t)$ is annealed after evidence warm-up.

In the second stage, the backbone is frozen and $g_\phi$ is trained on
matched pairs and semantically similar hard negatives. Let $b_i=1$
indicate mismatch and
$\widehat b_i=\sigma(U_{\mathrm{ale},i}^{\mathrm{raw}})$. The
class-weighted objective is
\begin{equation}
\mathcal L_{\mathrm{pair}}
=\sum_i \omega_{b_i}\operatorname{BCE}(\widehat b_i,b_i).
\label{eq:pair_loss}
\end{equation}

\paragraph{Epistemic objective.}
The sequence estimator first predicts masked token uncertainties from
observed positions. Following deep evidential regression
~\cite{amini2020deep}, the corresponding NIG marginal-likelihood loss is
\begin{equation}
\small
\mathcal L_{\mathrm{seq}}
=-\frac{1}{B}\sum_i
\log p\!\left(
\mathbf u_i^{\mathcal M_i}
\mid
\mathbf u_i^{\mathcal O_i};
\gamma_i^0,\nu_i^0,\alpha_i^0,\beta_i^0
\right).
\label{eq:seq_nig_loss}
\end{equation}
For semantic-error label $e_i$ and classifier logit $z_i$, binary
cross-entropy gives:
\begin{equation}
\mathcal L_{\mathrm{sem}}
=\sum_i \operatorname{BCE}(\sigma(z_i),e_i).
\label{eq:semantic_loss}
\end{equation}
To preserve relative error severity, let
$\mathcal P=\{(i,j):q_i>q_j+\delta\}$. The margin-ranking loss is
\begin{equation}
\small
\mathcal L_{\mathrm{rank}}
=\frac{1}{|\mathcal P|}
\sum_{(i,j)\in\mathcal P}
\max\left(
0,\,
m-U_{\mathrm{epi},i}^{\mathrm{raw}}
+U_{\mathrm{epi},j}^{\mathrm{raw}}
\right).
\label{eq:ranking_loss}
\end{equation}
The complete epistemic objective is
\begin{equation}
\mathcal L_{\mathrm{epi}}
=\lambda_{\mathrm{seq}}\mathcal L_{\mathrm{seq}}
+\lambda_{\mathrm{cal}}\mathcal L_{\mathrm{sem}}
+\lambda_{\mathrm{rank}}\mathcal L_{\mathrm{rank}}.
\label{eq:epi_loss}
\end{equation}
The three terms supervise masked token uncertainty, semantic-error
occurrence, and relative severity, respectively. We set
$(\lambda_{\mathrm{seq}},\lambda_{\mathrm{cal}},
\lambda_{\mathrm{rank}})=(0.02,0.05,5.0)$. Post-hoc calibration and
sequential fusion add no training objective.

\begin{table*}[!t]
\centering
\scriptsize
\setlength{\tabcolsep}{2.0pt}
\renewcommand{\arraystretch}{0.88}
\setlength{\aboverulesep}{0.3ex}
\setlength{\belowrulesep}{0.3ex}
\begin{tabular*}{\textwidth}{
@{\extracolsep{\fill}}lllccccc@{}}
\toprule
Dataset
& Aleatoric Method
& Epistemic Method
& AUROC$\uparrow$
& AUPRC$\uparrow$
& ECE$\downarrow$
& Brier$\downarrow$
& AURC$\downarrow$ \\
\midrule

\multirow{9}{*}{MINA}
& PCME++ & P(True)
& $0.678\pm0.016$ & $0.883\pm0.009$ & $0.058\pm0.016$
& $0.176\pm0.008$ & $0.670\pm0.009$ \\

& PCME++ & SelfCheckGPT
& $0.680\pm0.024$ & $0.883\pm0.011$
& $0.052\pm0.019$
& $0.175\pm0.009$ & $0.668\pm0.014$ \\

& UATVR & Semantic Entropy
& $0.710\pm0.025$ & $0.884\pm0.011$ & $0.057\pm0.009$
& $0.171\pm0.005$ & $0.644\pm0.024$ \\

& UATVR & SelfCheckGPT
& $0.711\pm0.024$ & $0.884\pm0.011$ & $0.052\pm0.011$
& $0.170\pm0.003$ & $0.644\pm0.024$ \\

& PAU & VL-Uncertainty
& $0.757\pm0.026$ & $0.912\pm0.010$ & $0.054\pm0.019$
& $0.161\pm0.007$ & $\underline{0.620\pm0.025}$ \\

& PAU & SelfCheckGPT
& $0.759\pm0.028$ & $0.912\pm0.010$ & $0.057\pm0.019$
& $0.161\pm0.007$ & $0.621\pm0.028$ \\

& FUME & Semantic Entropy
& $0.756\pm0.020$ & $0.915\pm0.008$ & $\underline{0.046\pm0.016}$
& $0.161\pm0.009$ & $0.626\pm0.017$ \\

& FUME & SelfCheckGPT
& $\underline{0.759\pm0.017}$
& $\underline{0.916\pm0.009}$
& $\mathbf{0.043\pm0.012}$
& $\underline{0.160\pm0.008}$
& $0.621\pm0.021$ \\

& \textbf{$U_{\mathrm{ale}}$ (Ours)}
& \textbf{$U_{\mathrm{epi}}$ (Ours)}
& $\mathbf{0.806\pm0.013}$
& $\mathbf{0.935\pm0.005}$
& $0.059\pm0.011$
& $\mathbf{0.150\pm0.007}$
& $\mathbf{0.596\pm0.018}$ \\

\midrule

\multirow{9}{*}{MIND}
& PCME++ & Semantic Entropy
& $0.621\pm0.027$ & $0.860\pm0.018$ & $0.067\pm0.021$
& $0.180\pm0.016$ & $0.715\pm0.032$ \\

& PCME++ & P(True)
& $0.624\pm0.020$ & $0.860\pm0.018$ & $0.084\pm0.028$
& $0.181\pm0.015$ & $0.708\pm0.019$ \\

& FUME & P(True)
& $0.655\pm0.022$ & $0.869\pm0.018$ & $0.091\pm0.033$
& $0.179\pm0.016$ & $0.690\pm0.027$ \\

& FUME & Semantic Entropy
& $0.656\pm0.029$ & $0.870\pm0.014$ & $0.079\pm0.023$
& $0.178\pm0.016$ & $0.691\pm0.029$ \\

& UATVR & P(True)
& $0.685\pm0.027$ & $0.890\pm0.013$ & $0.061\pm0.011$
& $0.168\pm0.008$ & $0.679\pm0.037$ \\

& UATVR & Semantic Entropy
& $0.686\pm0.021$ & $0.889\pm0.010$
& $\mathbf{0.058\pm0.015}$
& $0.169\pm0.009$ & $0.680\pm0.033$ \\

& PAU & Semantic Entropy
& $0.712\pm0.014$ & $0.898\pm0.008$
& $\underline{0.059\pm0.010}$
& $0.166\pm0.012$ & $0.662\pm0.028$ \\

& PAU & P(True)
& $\underline{0.721\pm0.020}$
& $\underline{0.900\pm0.013}$
& $0.066\pm0.020$
& $\underline{0.164\pm0.010}$
& $\underline{0.657\pm0.028}$ \\

& \textbf{$U_{\mathrm{ale}}$ (Ours)}
& \textbf{$U_{\mathrm{epi}}$ (Ours)}
& $\mathbf{0.755\pm0.027}$
& $\mathbf{0.923\pm0.003}$
& $0.061\pm0.014$
& $\mathbf{0.158\pm0.007}$
& $\mathbf{0.647\pm0.045}$ \\

\bottomrule
\end{tabular*}
\caption{Overall reliability under a 50\% input-mismatch setting, averaged over five mixture seeds.}
\label{tab:total_fusion}
\end{table*}

\section{Experiments}
\label{sec:experiments}

\subsection{Experimental Setup}

\noindent\textbf{Datasets and task construction.}
We conduct experiments on two multimodal intention-analysis datasets: MINA~\cite{lu2026mina}, which contains social-media video-text-audio modal information, and MIND~\cite{xu2024mind}, which consists of shopping-oriented multimodal product pairs.
Qwen3-VL-8B-Instruct is adopted as the backbone model for both datasets.
To simulate challenging input-level inconsistencies, we construct hard mismatches by pairing each visual input with semantically similar but
incorrect text.
Output-level semantic error is measured by $1-\operatorname{BERTScore}$, and its binary decision threshold is determined using the validation set.
A sample is considered unreliable if it contains an input mismatch or, when aligned, incurs a semantic output error.

\noindent\textbf{Baselines.}
For input-level uncertainty estimation, we compare DUE with PCME++~\cite{chun2024improved}, UATVR~\cite{fang2023uatvr}, PAU~\cite{li2023prototype}, and FUME~\cite{duan2025fuzzy}.
For output-level uncertainty estimation, the baselines include token-entropy statistics~\cite{malinin2021uncertainty},SelfCheckGPT~\cite{manakul2023selfcheckgpt}, Semantic Entropy~\cite{kuhn2023semantic}, P(True)~\cite{kadavath2022language}, and
VL-Uncertainty~\cite{zhang2024vluncertainty}.
For a fair comparison, all methods use identical data partitions, reliability
labels, and calibration sets.

\noindent\textbf{Evaluation metrics.}
We evaluate uncertainty estimation using AUROC~\cite{fawcett2006roc}, AUPRC~\cite{saito2015precision},
Spearman correlation~\cite{spearman1904proof}, expected calibration error (ECE)~\cite{guo2017calibration},
Brier score~\cite{brier1950verification}, and area under the risk--coverage curve (AURC)~\cite{ding2020revisiting}.
Arrows indicate whether higher or lower values are preferred.
The best and second-best results are highlighted in bold and underlined, respectively. All calibrators are fitted exclusively on the calibration set, while the epistemic calibrator is fitted only on aligned calibration samples.

\noindent\textbf{Implementation details.}
All experiments are conducted on NVIDIA Tesla V100 GPUs.
DUE aggregates $K=5$ intermediate layers and uses $S=10$ stochastic samples.
Its sequence encoder has a hidden dimension of 96 and a dropout rate of 0.1.
The input-level uncertainty estimator is trained for 10 epochs with a batch size of 16 and a learning rate of $10^{-3}$.
Additional details regarding data splits, preprocessing, hyperparameter settings, baseline implementations, theoretical analysis, complete results, confidence intervals, computational efficiency, and failure cases are
provided in the supplementary materials.

\subsection{Main Results}\label{sec:overall_reliability}

Table~\ref{tab:total_fusion} evaluates overall unreliability on test sets with 50\% mismatched inputs. DUE achieves the highest AUROC and AUPRC and the lowest Brier score and AURC on both datasets.
In terms of mean AUROC, it improves the strongest baseline fusion from 0.759 to 0.806 on MINA and from 0.721 to 0.755 on MIND. These results indicate better discrimination of unreliable samples, probability estimation, and risk ranking. Although DUE does not obtain the lowest ECE, its calibration error remains comparable to the baselines.

\paragraph{Input-side uncertainty.}
Table~\ref{tab:input_uncertainty} evaluates cross-modal mismatch detection.
$U_{\mathrm{ale}}$ achieves the best AUROC, AUPRC, Brier score, and AURC on both datasets, improving AUROC over the strongest baseline by 3.13 points on MINA and 5.91 points on MIND.
The results show that the proposed estimator more effectively identifies semantically similar but incorrectly paired multimodal inputs. Its ECE is competitive on MIND but remains suboptimal on MINA.

\begin{table}[!htbp]
\centering
\scriptsize
\setlength{\tabcolsep}{0.8pt}
\renewcommand{\arraystretch}{0.88}
\setlength{\aboverulesep}{0.3ex}
\setlength{\belowrulesep}{0.3ex}
\begin{tabular*}{\columnwidth}{
@{\extracolsep{\fill}}llccccc@{}}
\toprule
Dataset & Method
& AUROC$\uparrow$
& AUPRC$\uparrow$
& ECE$\downarrow$
& Brier$\downarrow$
& AURC$\downarrow$ \\
\midrule
\multirow{5}{*}{MINA}
& PCME++ & $0.8028$ & $0.8187$ & $\mathbf{0.0438}$ & $0.1847$ & $0.3009$ \\
& UATVR & $0.8337$ & $0.8270$ & $\underline{0.0566}$ & $0.1679$ & $0.2775$ \\
& PAU & $\underline{0.8922}$ & $0.8718$ & $0.0945$
& $\underline{0.1387}$ & $\underline{0.2403}$ \\
& FUME & $0.8780$ & $\underline{0.8895}$ & $0.0699$ & $0.1517$ & $0.2497$ \\
& $U_{\mathrm{ale}}$ (Ours)
& $\mathbf{0.9235}$ & $\mathbf{0.9340}$ & $0.0868$
& $\mathbf{0.1223}$ & $\mathbf{0.2298}$ \\
\midrule
\multirow{5}{*}{MIND}
& PCME++ & $0.7283$ & $0.7511$ & $0.0419$ & $0.2141$ & $0.3681$ \\
& UATVR & $0.8053$ & $0.8105$ & $0.0453$ & $0.1830$ & $0.3048$ \\
& PAU & $\underline{0.8351}$ & $\underline{0.8433}$ & $\underline{0.0382}$
& $\underline{0.1674}$ & $\underline{0.2816}$ \\
& FUME & $0.8083$ & $0.7995$ & $\mathbf{0.0369}$ & $0.1787$ & $0.2920$ \\
& $U_{\mathrm{ale}}$ (Ours)
& $\mathbf{0.8942}$ & $\mathbf{0.9020}$ & $\mathbf{0.0369}$
& $\mathbf{0.1365}$ & $\mathbf{0.2382}$ \\
\bottomrule
\end{tabular*}
\caption{Input-side aleatoric uncertainty for cross-modal mismatch
detection on MINA and MIND.}
\label{tab:input_uncertainty}
\end{table}

\paragraph{Model-side uncertainty.}
Table~\ref{tab:output_uncertainty} evaluates semantic-error detection on
aligned inputs. $U_{\mathrm{epi}}$ achieves the best AUROC, AUPRC,
Spearman correlation, and AURC on both datasets. The improvement is more
pronounced on MINA, whereas the modest AUROC gain on MIND indicates that
semantic-error detection remains challenging. Moreover, DUE matches the
cost of one-pass entropy baselines and is substantially more efficient
than methods requiring repeated generation. Its ECE is not the best,
leaving room for improved calibration.

\begin{table}[!htbp]
\centering
\scriptsize
\setlength{\tabcolsep}{0.4pt}
\renewcommand{\arraystretch}{0.88}
\setlength{\aboverulesep}{0.3ex}
\setlength{\belowrulesep}{0.3ex}
\begin{tabular*}{\columnwidth}{
@{\extracolsep{\fill}}llcccccc@{}}
\toprule
Dataset & Method
& AUROC$\uparrow$
& AUPRC$\uparrow$
& $\rho\uparrow$
& ECE$\downarrow$
& AURC$\downarrow$
& Time$\downarrow$ \\
\midrule
\multirow{7}{*}{MINA}
& Entropy Mean
& $\underline{0.578}$ & $\underline{0.598}$ & $\underline{0.163}$
& $0.036$ & $\underline{0.476}$ & $\mathbf{1.0\times}$ \\
& Entropy Std
& $0.561$ & $0.596$ & $0.120$ & $\mathbf{0.012}$ & $0.488$
& $\mathbf{1.0\times}$ \\
& SelfCheckGPT
& $0.509$ & $0.532$ & $0.037$ & $0.037$ & $0.512$ & $6.0\times$ \\
& Semantic Entropy
& $0.496$ & $0.528$ & $0.012$ & $0.041$ & $0.527$ & $6.0\times$ \\
& P(True)
& $0.506$ & $0.538$ & $-0.079$ & $0.019$ & $0.516$
& $\underline{1.9\times}$ \\
& VL-Uncertainty
& $0.507$ & $0.528$ & $0.026$ & $\underline{0.018}$ & $0.512$ & $5.0\times$ \\
& $U_{\mathrm{epi}}$ (Ours)
& $\mathbf{0.631}$ & $\mathbf{0.648}$ & $\mathbf{0.265}$
& $0.049$ & $\mathbf{0.440}$ & $\mathbf{1.0\times}$ \\
\midrule
\multirow{7}{*}{MIND}
& Entropy Mean
& $0.514$ & $0.507$ & $0.027$ & $0.042$ & $0.493$ & $\mathbf{1.0\times}$ \\
& Entropy Std
& $0.496$ & $0.494$ & $0.010$ & $0.050$ & $0.500$ & $\mathbf{1.0\times}$ \\
& SelfCheckGPT
& $\underline{0.538}$ & $\underline{0.541}$ & $\underline{0.061}$
& $\mathbf{0.010}$ & $0.487$ & $6.0\times$ \\
& Semantic Entropy
& $0.525$ & $\underline{0.541}$ & $0.039$ & $0.102$ & $0.487$ & $6.0\times$ \\
& P(True)
& $0.534$ & $0.527$ & $0.048$ & $0.040$ & $0.485$
& $\underline{1.9\times}$ \\
& VL-Uncertainty
& $0.531$ & $0.539$ & $0.040$ & $\underline{0.013}$
& $\underline{0.481}$ & $5.0\times$ \\
& $U_{\mathrm{epi}}$ (Ours)
& $\mathbf{0.548}$ & $\mathbf{0.551}$ & $\mathbf{0.099}$
& $0.039$ & $\mathbf{0.469}$ & $\mathbf{1.0\times}$ \\
\bottomrule
\end{tabular*}
\caption{Sentence-level epistemic uncertainty and relative inference cost;
$\rho$ denotes Spearman correlation.}
\label{tab:output_uncertainty}
\end{table}

\subsection{Selective Prediction Analysis}

\begin{figure}[h]
    \centering
    \includegraphics[
        width=0.95\columnwidth,
        height=4.2cm,
        keepaspectratio
    ]{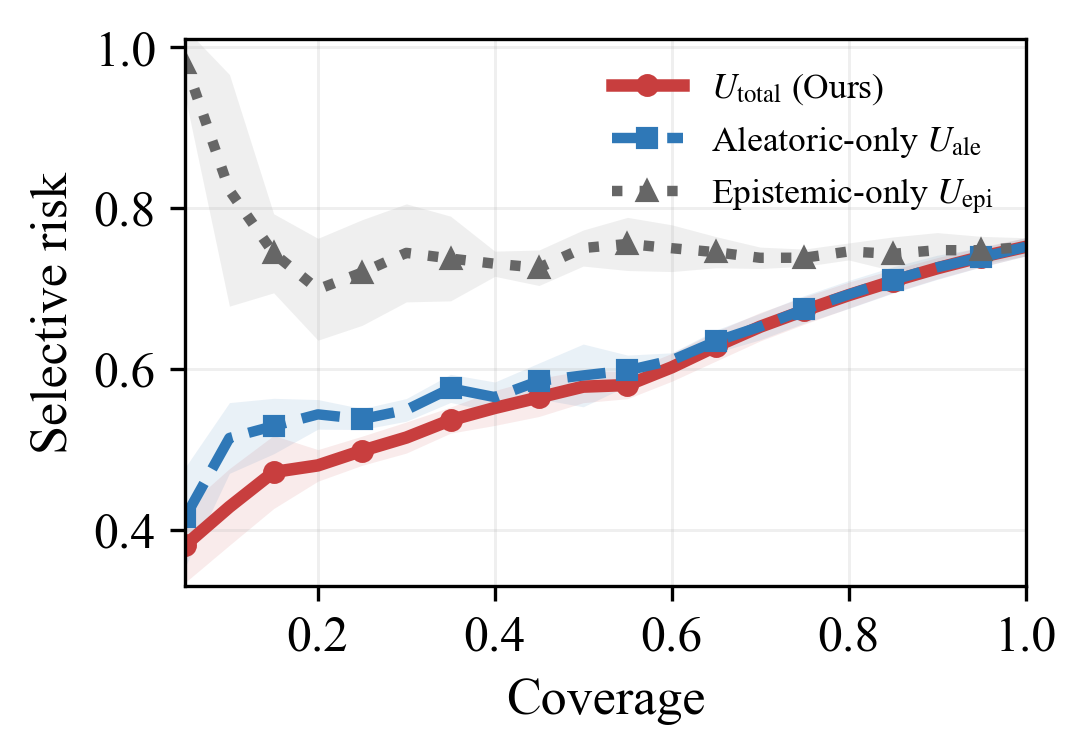}
    \caption{Risk--coverage on MINA at 50\% mismatch; shading denotes
    standard deviation over five seeds.}
    \label{fig:risk_coverage}
\end{figure}

Figure~\ref{fig:risk_coverage} evaluates selective prediction by progressively retaining samples with lower estimated risk.
$U_{\mathrm{total}}$ consistently achieves lower selective risk than either branch, particularly at low-to-medium coverage. This suggests $U_{\mathrm{ale}}$ and $U_{\mathrm{epi}}$ capture complementary failure signals, enabling their fusion to rank unreliable samples more accurately and reject them earlier. At higher coverage, curves converge as more high-risk samples are retained.

\begin{table*}[!t]
\centering
\scriptsize
\setlength{\tabcolsep}{2.6pt}
\renewcommand{\arraystretch}{0.88}
\setlength{\aboverulesep}{0.3ex}
\setlength{\belowrulesep}{0.3ex}
\begin{tabular*}{\textwidth}{
@{\extracolsep{\fill}}llccccc@{}}
\toprule
Analysis & Method
& AUROC$\uparrow$
& AUPRC$\uparrow$
& ECE$\downarrow$
& Brier$\downarrow$
& AURC$\downarrow$ \\
\midrule
\multirow{3}{*}{Decomposition}
& Epistemic-only $U_{\mathrm{epi}}$
& $0.588\pm0.009$ & $0.844\pm0.007$ & $\mathbf{0.030\pm0.008}$
& $0.184\pm0.004$ & $0.708\pm0.012$ \\
& Aleatoric-only $U_{\mathrm{ale}}$
& $\underline{0.790\pm0.008}$ & $\mathbf{0.937\pm0.004}$
& $0.084\pm0.014$ & $\underline{0.159\pm0.008}$
& $\underline{0.607\pm0.011}$ \\
& Full $U_{\mathrm{total}}$
& $\mathbf{0.806\pm0.011}$ & $\underline{0.935\pm0.005}$
& $\underline{0.059\pm0.010}$ & $\mathbf{0.150\pm0.006}$
& $\mathbf{0.596\pm0.016}$ \\
\midrule
\multirow{5}{*}{Fusion}
& Calibrated Average
& $0.799\pm0.013$ & $0.930\pm0.006$ & $0.264\pm0.009$
& $0.221\pm0.009$ & $0.600\pm0.017$ \\
& Separate Calibration + Weighted Sum
& $0.801\pm0.012$ & $0.933\pm0.006$ & $0.260\pm0.011$
& $0.227\pm0.013$ & $0.600\pm0.016$ \\
& Noisy-OR w/o Conditional Calibration
& $\underline{0.804\pm0.011}$ & $0.935\pm0.004$ & $0.120\pm0.010$
& $0.169\pm0.006$ & $\underline{0.597\pm0.016}$ \\
& Shared Calibration + Noisy-OR
& $0.803\pm0.010$ & $\mathbf{0.935\pm0.004}$
& $\mathbf{0.049\pm0.007}$ & $\mathbf{0.149\pm0.004}$
& $0.598\pm0.016$ \\
& Sequential Noisy-OR (Ours)
& $\mathbf{0.806\pm0.011}$ & $\underline{0.935\pm0.005}$
& $\underline{0.059\pm0.010}$ & $\underline{0.150\pm0.006}$
& $\mathbf{0.596\pm0.016}$ \\
\bottomrule
\end{tabular*}
\caption{Decomposition and fusion analysis on MINA with 50\% mismatch,
averaged over five seeds.}
\label{tab:fusion_analysis}
\end{table*}

\subsection{Sensitivity to Input-Mismatch Prevalence}
\begin{figure}[h]
\centering
\includegraphics[width=0.98\columnwidth]
{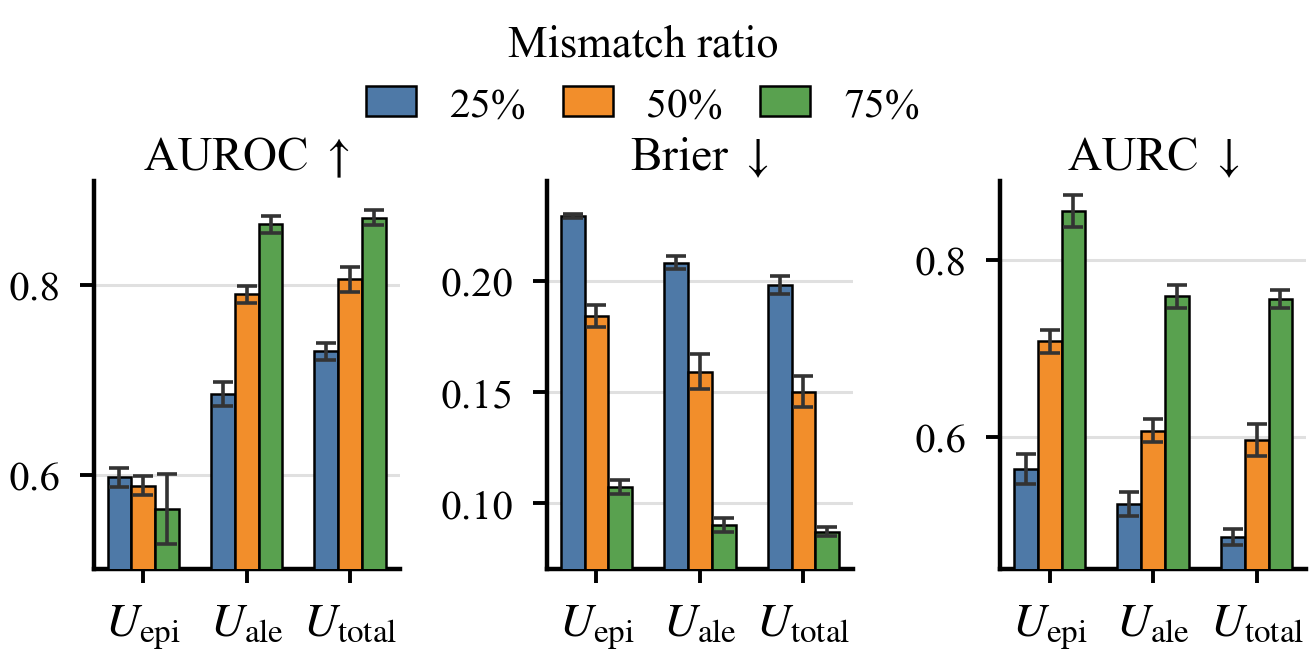}
\caption{Input-mismatch sensitivity on MINA (mean and standard deviation over five seeds).}
\label{fig:mismatch_sensitivity}
\end{figure}

Figure~\ref{fig:mismatch_sensitivity} evaluates mismatch ratios from 25\% to 75\%.
As mismatch becomes more prevalent, the AUROC of $U_{\mathrm{epi}}$ approaches chance level, whereas $U_{\mathrm{ale}}$ improves substantially, showing that input-side uncertainty directly captures cross-modal mismatch.
$U_{\mathrm{total}}$ remains the strongest across all three metrics, preserving mismatch detection while incorporating output-side risk.
The increase in AURC mainly reflects the higher prevalence of unreliable samples.
Besides, Brier scores are likewise prevalence-dependent.

\subsection{Ablation and Fusion Strategy Analysis}
\label{sec:ablation}
\paragraph{Uncertainty decomposition and fusion.}
As shown in Table~\ref{tab:fusion_analysis}, combining the two branches
improves AUROC from 0.790 to 0.806 and reduces Brier score and AURC,
indicating complementary input- and model-side risk signals. Conditional
calibration further lowers ECE from 0.120 to 0.059 and Brier score from
0.169 to 0.150. Sequential Noisy-OR improves discrimination and ranking,
whereas shared calibration slightly improves probability calibration.

\begin{table}[h]
\centering
\scriptsize
\setlength{\tabcolsep}{2.0pt}
\renewcommand{\arraystretch}{0.88}
\setlength{\aboverulesep}{0.3ex}
\setlength{\belowrulesep}{0.3ex}
\resizebox{\columnwidth}{!}{%
\begin{tabular}{lcccccc}
\toprule
Variant
& AUROC$\uparrow$
& AUPRC$\uparrow$
& Spearman$\uparrow$
& ECE$\downarrow$
& Brier$\downarrow$
& AURC$\downarrow$ \\
\midrule
Full
& $\mathbf{0.676}$ & $\mathbf{0.670}$ & $\mathbf{0.362}$
& $0.065$ & $\mathbf{0.229}$ & $\mathbf{0.384}$ \\
w/o Attention
& $0.669$ & $0.657$ & $\underline{0.353}$
& $\mathbf{0.049}$ & $0.231$ & $0.390$ \\
w/o Statistical Features
& $\underline{0.670}$ & $\underline{0.665}$ & $0.347$
& $\underline{0.049}$ & $\underline{0.230}$ & $\underline{0.389}$ \\
w/o Sequence Loss
& $0.658$ & $0.648$ & $0.324$
& $0.052$ & $0.236$ & $0.401$ \\
w/o Ranking Loss
& $0.638$ & $0.621$ & $0.284$
& $0.056$ & $0.239$ & $0.410$ \\
\bottomrule
\end{tabular}%
}
\caption{Fixed-seed epistemic uncertainty ablation study.}
\label{tab:uep_ablation}
\end{table}

\paragraph{Epistemic uncertainty ablation.}
As shown in Table~\ref{tab:uep_ablation}, removing the sequence or ranking
loss decreases AUROC from 0.676 to 0.658 and 0.638, respectively, with
similar declines in AUPRC, Spearman correlation, and AURC.
This indicates that both sequential modeling and severity-aware ranking contribute to risk estimation, with the ranking loss having the larger effect.
Removing attention or statistical features causes smaller but consistent degradation in discrimination and ranking, suggesting local sequence cues and global statistics provide complementary information.
\subsection{Case Study}
\label{sec:case_study}

\begin{figure}[h]
    \centering
    \includegraphics[width=\columnwidth]{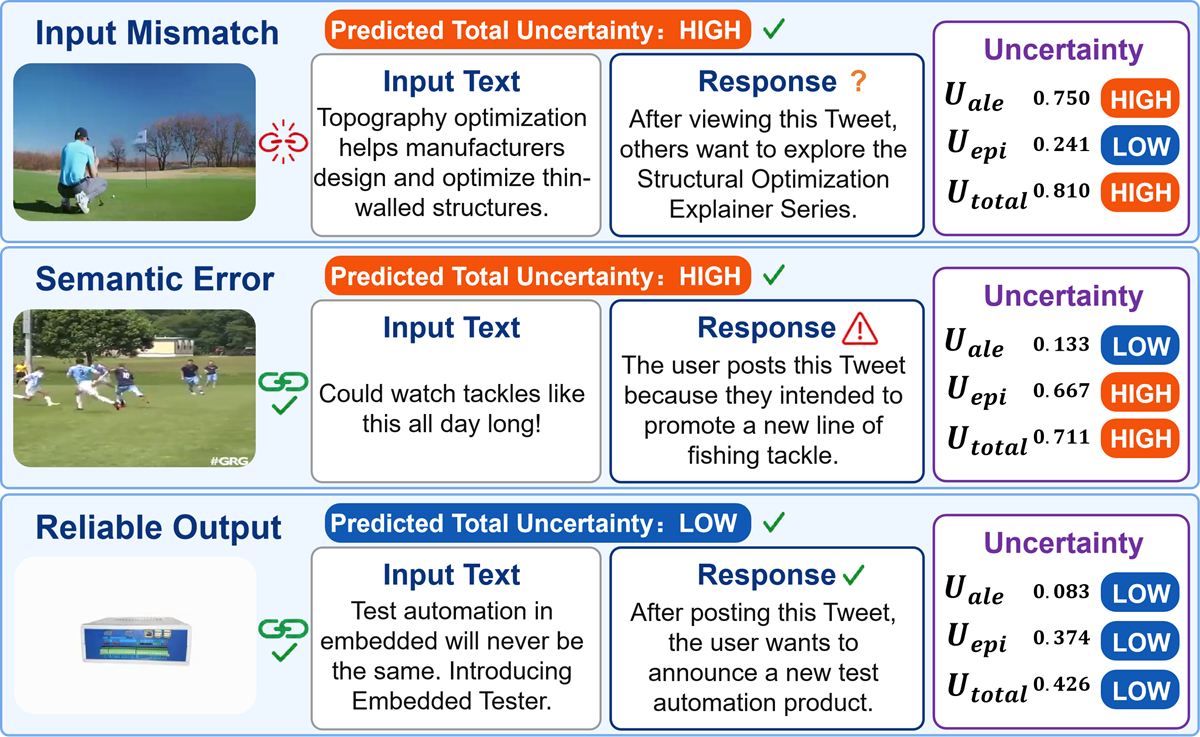}
    \caption{Representative cases: input mismatch via $U_{\mathrm{ale}}$,
semantic error via $U_{\mathrm{epi}}$, and fused risk via
$U_{\mathrm{total}}$.}
    \label{fig:case_study}
\end{figure}

Figure~\ref{fig:case_study} illustrates source-specific uncertainty. The
mismatched input yields high $U_{\mathrm{ale}}$ and
$U_{\mathrm{total}}$ despite low $U_{\mathrm{epi}}$, whereas the aligned
semantic error yields low $U_{\mathrm{ale}}$ but high
$U_{\mathrm{epi}}$. All scores remain lower for the reliable case. These
examples demonstrate complementary risk attribution and unified
reliability estimation through fusion.

\section{Conclusion}
We presented DUE, a dual-side uncertainty framework for reliable multimodal intention analysis.
DUE estimates sentence-level epistemic uncertainty via sequence-aware NIG modeling of token uncertainty within a single generation and input-side aleatoric uncertainty via evidential multimodal matching.
Event-specific calibration and sequential fusion yield an interpretable overall failure probability.
Experiments conducted on the MINA and MIND datasets demonstrate improved input-mismatch detection, semantic-error detection, and overall reliability estimation without requiring repeated generation.

\section*{Limitations}
DUE jointly estimates input-side and model-side risks from a single
generation, with experiments on MINA and MIND supporting its effectiveness.
The current study leaves two directions for future work:

\begin{itemize}
    \item Due to limited computational resources, our evaluation
    currently uses Qwen3-VL-8B-Instruct. Future work will extend
    the evaluation to larger models to further examine DUE's
    applicability across model scales.

    \item We plan to build an uncertainty quantification benchmark from the existing samples in future work, with refined quality annotations and standardized evaluation protocols, to support reproducible comparisons and facilitate research by the broader community.
\end{itemize}

\section*{Ethical Considerations}
DUE aims to improve the reliability of multimodal intention analysis
by estimating input-side and model-side risks. Its practical deployment
should nevertheless consider potential social risks.
First, inferring intentions from multimodal behavioral observations
may reveal sensitive preferences or personal attributes. Such
applications should respect user consent, limit data collection,
and restrict access to sensitive information.
Second, biases in the underlying data, generation model, or quality
supervision may affect risk estimates across user groups and intention
categories. Deployment should therefore include context-appropriate
fairness assessments and monitoring.
Finally, uncertainty estimates should not be treated as guarantees
of correctness or used as the sole basis for consequential decisions
about individuals. Appropriate data governance, transparent
communication of the estimates, and human oversight are important
to reduce misuse and support responsible deployment.

\bibliography{references,supplement_references}

\ifincludeappendix
  \appendix
  \clearpage

\section{Overview of the Appendix}
This appendix provides the theoretical analysis, data and
preprocessing details, implementation settings, complete quantitative
results, and additional robustness analyses for DUE. We use
$U_{\mathrm{ale}}^{\mathrm{raw}}$ and
$U_{\mathrm{epi}}^{\mathrm{raw}}$ for the uncalibrated input-side
aleatoric and sentence-level epistemic scores, respectively. Their
calibrated event probabilities are denoted by
$\widetilde U_{\mathrm{ale}}$ and $\widetilde U_{\mathrm{epi}}$, while
$U_{\mathrm{total}}$ is the overall failure probability produced by
sequential fusion.

\section{Theoretical Justification of Single-Generation Hidden-State Sampling}
\label{app:hidden_sampling}

We provide a theoretical justification for constructing an ordered token
uncertainty sequence from the cached hidden states of a single
autoregressive response. Our analysis does not assume that representations
from different layers are exact samples from the parameter posterior.
Instead, it shows that high-layer mixing induces a controlled local
predictive distribution and that the resulting token uncertainty is a
consistent Monte Carlo estimate of this distribution.

\paragraph{Setup.}
For generation position $t$, let
$\mathcal C_t=(X,Y_{<t})$ denote the input and the generated prefix, and let
$\{h_t^{(k)}\}_{k\in\mathcal L}$ be the cached hidden states from a selected
set $\mathcal L$ of $K$ high layers. Independently for each token, we draw
$\mathbf w_t^{(s)}\sim\operatorname{Dir}(\mathbf a)$ and define
\begin{equation}
\widetilde h_t^{(s)}
=
\sum_{k\in\mathcal L}w_{tk}^{(s)}h_t^{(k)},
\qquad
p_t^{(s)}
=
g(\widetilde h_t^{(s)}),
\label{eq:app_internal_sample}
\end{equation}
where the frozen output mapping is
\begin{equation}
g(h)=
\operatorname{Softmax}\!\left(
\operatorname{LMHead}(\operatorname{FinalNorm}(h))
\right).
\end{equation}
The Monte Carlo predictive distribution and token uncertainty are
\begin{equation}
\bar p_t^{(S)}
=
\frac{1}{S}\sum_{s=1}^{S}p_t^{(s)},
\qquad
u_t^{(S)}
=
H(\bar p_t^{(S)}).
\label{eq:app_token_uncertainty}
\end{equation}

\paragraph{Proposition A.1 (Information retained by high-layer
representations).}
Let $Z_t$ denote the random representation induced by the internal mixing
in Eq.~\eqref{eq:app_internal_sample}, and let $Y_t$ be the next-token
random variable. Suppose that the selected high-layer representation is
approximately sufficient for the output decision, i.e.,
\begin{equation}
H(Y_t\mid Z_t,\mathcal C_t)\leq\delta_t,
\label{eq:app_sufficiency}
\end{equation}
where $\delta_t$ is small. Then
\begin{equation}
I(Y_t;Z_t\mid\mathcal C_t)
\geq
H(Y_t\mid\mathcal C_t)-\delta_t.
\label{eq:app_information_bound}
\end{equation}

\paragraph{Proof.}
By the definition of conditional mutual information,
\begin{align}
I(Y_t;Z_t\mid\mathcal C_t)
&=
H(Y_t\mid\mathcal C_t)
-
H(Y_t\mid Z_t,\mathcal C_t) \nonumber\\
&\geq
H(Y_t\mid\mathcal C_t)-\delta_t,
\end{align}
where the inequality follows from
Eq.~\eqref{eq:app_sufficiency}.
Therefore, except for a residual term bounded by $\delta_t$, the
uncertainty of the next-token decision is retained in the sampled
high-layer representation. \hfill$\square$

\paragraph{Proposition A.2 (Controlled representation-space sampling).}
Assume that the selected high-layer states satisfy
\begin{equation}
\begin{aligned}
\cos(h_t^{(i)},h_t^{(j)})
&\geq 1-\epsilon,\\
\|h_t^{(i)}\|_2
&\leq M,\\
\left|
\|h_t^{(i)}\|_2-\|h_t^{(j)}\|_2
\right|
&\leq \rho.
\end{aligned}
\label{eq:app_layer_conditions}
\end{equation}
for all $i,j\in\mathcal L$. Define
\begin{equation}
r=\sqrt{\rho^2+2\epsilon M^2}.
\end{equation}
Then every Dirichlet mixture remains within distance $r$ of any selected
layer state. Moreover, if $g$ is $L_g$-Lipschitz continuous on the convex
hull of the selected states, then any two internally sampled predictive
distributions satisfy
\begin{equation}
\left\|
p_t^{(s)}-p_t^{(s')}
\right\|_2
\leq
2L_g r.
\label{eq:app_predictive_bound}
\end{equation}

\paragraph{Proof.}
For any $i,j\in\mathcal L$,
\begin{align}
\|h_t^{(i)}-h_t^{(j)}\|_2^2
&=
\left(
\|h_t^{(i)}\|_2-\|h_t^{(j)}\|_2
\right)^2 \nonumber\\
&\quad+
2\|h_t^{(i)}\|_2\|h_t^{(j)}\|_2 \nonumber\\
&\qquad{}\times\left(
1-\cos(h_t^{(i)},h_t^{(j)})
\right) \nonumber\\
&\leq
\rho^2+2\epsilon M^2
=
r^2.
\label{eq:app_hidden_distance}
\end{align}
Because Dirichlet weights are nonnegative and sum to one,
$\widetilde h_t^{(s)}$ is a convex combination of the selected layer
states. For any reference layer $j\in\mathcal L$,
\begin{align}
\left\|
\widetilde h_t^{(s)}-h_t^{(j)}
\right\|_2
&=
\left\|
\sum_{k\in\mathcal L}
w_{tk}^{(s)}
\left(
h_t^{(k)}-h_t^{(j)}
\right)
\right\|_2 \nonumber\\
&\leq
\sum_{k\in\mathcal L}
w_{tk}^{(s)}
\left\|
h_t^{(k)}-h_t^{(j)}
\right\|_2
\leq r.
\end{align}
Consequently, for two samples $s$ and $s'$,
\begin{equation}
\left\|
\widetilde h_t^{(s)}
-
\widetilde h_t^{(s')}
\right\|_2
\leq 2r.
\end{equation}
Applying the Lipschitz continuity of $g$ gives
\begin{align}
\left\|
p_t^{(s)}-p_t^{(s')}
\right\|_2
&\leq
L_g
\left\|
\widetilde h_t^{(s)}
-
\widetilde h_t^{(s')}
\right\|_2
\leq2L_g r.
\end{align}
Thus, internal mixing produces controlled perturbations around a shared
high-level semantic state instead of arbitrary representations.
\hfill$\square$

For normalized hidden states, $\rho=0$ and $M=1$, and the radius reduces
to $r=\sqrt{2\epsilon}$. The Dirichlet concentration further controls the
sampling strength. For a symmetric distribution
$\operatorname{Dir}(\alpha\mathbf 1)$,
\begin{equation}
\mathbb E[w_k]=\frac{1}{K},
\qquad
\operatorname{Var}(w_k)
=
\frac{K-1}{K^2(K\alpha+1)}.
\label{eq:app_dirichlet_variance}
\end{equation}
Hence, increasing $\alpha$ reduces the variance of the internal samples,
whereas a smaller $\alpha$ exposes stronger layer-wise disagreement.

\paragraph{Proposition A.3 (Consistency of token uncertainty).}
Let the Dirichlet weights be independently sampled and define
\begin{equation}
p_t^\star
=
\mathbb E_{\mathbf w}
\left[
g\left(
\sum_{k\in\mathcal L}w_kh_t^{(k)}
\right)
\right].
\end{equation}
Then
\begin{equation}
\bar p_t^{(S)}
\xrightarrow[S\rightarrow\infty]{\mathrm{a.s.}}
p_t^\star,
\qquad
u_t^{(S)}
\xrightarrow[S\rightarrow\infty]{\mathrm{a.s.}}
H(p_t^\star).
\label{eq:app_consistency}
\end{equation}

\paragraph{Proof.}
Each $p_t^{(s)}$ is an independent and bounded random vector on the
probability simplex. The strong law of large numbers therefore gives
\begin{equation}
\frac{1}{S}\sum_{s=1}^{S}p_t^{(s)}
\xrightarrow[S\rightarrow\infty]{\mathrm{a.s.}}
\mathbb E_{\mathbf w}[p_t^{(\mathbf w)}]
=
p_t^\star.
\end{equation}
Since Shannon entropy is continuous on a finite-dimensional probability
simplex, the continuous mapping theorem yields
\begin{equation}
H(\bar p_t^{(S)})
\xrightarrow[S\rightarrow\infty]{\mathrm{a.s.}}
H(p_t^\star).
\end{equation}
Therefore, $u_t^{(S)}$ is a consistent Monte Carlo estimate of the
predictive entropy induced by high-layer mixing. \hfill$\square$

A finite-sample result follows from Hoeffding's inequality. For a
vocabulary of size $|\mathcal V|$ and any $\xi>0$,
\begin{equation}
\Pr\left(
\|\bar p_t^{(S)}-p_t^\star\|_\infty\geq\xi
\right)
\leq
2|\mathcal V|\exp(-2S\xi^2).
\label{eq:app_finite_sample}
\end{equation}
Thus, the internal predictive estimate becomes increasingly stable as
the number of hidden-state samples grows.

\paragraph{Ordered uncertainty sequence.}
For a single generated response
$Y=(Y_1,\ldots,Y_T)$, the above procedure is applied at each position
under its actual autoregressive condition
$\mathcal C_t=(X,Y_{<t})$. This produces
\begin{equation}
\mathbf u^{(S)}
=
\left(
u_1^{(S)},u_2^{(S)},\ldots,u_T^{(S)}
\right).
\label{eq:app_ordered_sequence}
\end{equation}
The sequence preserves the generation order because every $u_t^{(S)}$
is indexed by its position and conditioned on the preceding tokens.
No independence assumption is imposed across positions. Consequently,
local peaks, transitions, and dependencies remain available to the
subsequent sequence-conditioned NIG estimator.

\paragraph{Discussion.}
Here, ``single generation'' means that only one autoregressive response
is decoded. The $S$ internal samples reuse its cached high-layer states
and require no repeated response generation. Propositions A.1--A.3 do
not claim that layer mixtures are exact samples from the parameter
posterior. They establish that the proposed procedure constructs a
controlled representation-space sampling distribution and consistently
estimates its token-level predictive uncertainty.

\section{Detailed Experimental Setup}
\label{app:experimental_setup}

\subsection{Datasets and Source-Level Splits}
\label{app:data_split}

We evaluate on MINA~\cite{lu2026mina} and
MIND~\cite{xu2024mind}. MINA contains multimodal social-media
observations and natural-language intention descriptions, whereas MIND
focuses on multimodal shopping-intention understanding. In both datasets,
the visual content and textual context form the multimodal input, and the
target is an open-ended natural-language intention. The official dataset
and code entry points are
\url{https://github.com/omeroyangtao/MINA/tree/main} for MINA and
\url{https://github.com/HKUST-KnowComp/MIND_Distillation/tree/main} for
MIND. Both datasets are used under the access conditions specified by
their respective repositories.

All partitions are constructed at the source level. Samples originating
from the same post, video, or product instance are assigned to only one
partition. The training partition optimizes the learned uncertainty
estimators, the validation partition is used for model selection and for
fixing the semantic-error threshold, and a disjoint calibration partition
fits the post-hoc mappings. Test samples are evaluated without additional
parameter selection. This source-level restriction is also applied when
constructing negative pairs, preventing the visual or textual source of a
test example from entering training or calibration through a mismatched
pair.

Table~\ref{tab:app_source_splits} gives the exact source counts used in
the final experiments. For MINA, each video is uniformly sampled into 16
frames before source-level partitioning. The split is generated with seed
42 from 1,209 common sources: 967 sources form the outer training pool and
are further divided into training and validation sets, while 120 and 122
disjoint sources are reserved for calibration and testing. For MIND, we
retain records whose \texttt{agreement} field is \texttt{yes} and use the
first 3,000 records in repository order. These records are then partitioned
using the same source-level protocol.

\begin{table}[t]
\centering
\small
\setlength{\tabcolsep}{6pt}
\begin{tabular}{lrrrr}
\toprule
Dataset & Train & Validation & Calibration & Test \\
\midrule
MINA & $773$ & $194$ & $120$ & $122$ \\
MIND & $1{,}920$ & $480$ & $300$ & $300$ \\
\bottomrule
\end{tabular}
\caption{Source-level partitions used in the final experiments.}
\label{tab:app_source_splits}
\end{table}

For mismatch evaluation, every calibration and test source contributes
one matched and one mismatched pair. Table~\ref{tab:app_pair_splits}
reports the resulting pair counts. Matched and mismatched pairs are
constructed only within their assigned partition.

\begin{table}[t]
\centering
\small
\setlength{\tabcolsep}{2.6pt}
\begin{tabular}{lrrrr}
\toprule
& \multicolumn{2}{c}{Calibration} &
\multicolumn{2}{c}{Test} \\
\cmidrule(lr){2-3}\cmidrule(lr){4-5}
Dataset & Matched & Mismatched & Matched & Mismatched \\
\midrule
MINA & $120$ & $120$ & $122$ & $122$ \\
MIND & $300$ & $300$ & $300$ & $300$ \\
\bottomrule
\end{tabular}
\caption{Matched and mismatched pair counts for calibration and testing.}
\label{tab:app_pair_splits}
\end{table}

\subsection{Construction of Input Mismatches}
\label{app:mismatch_construction}

Original visual--textual pairs are treated as matched examples. Mismatched
examples are constructed by pairing each visual input with a textual
context from another source in the same partition. The primary protocol
uses semantically similar but incorrect contexts, producing hard negatives
that cannot be identified from coarse topic differences alone.

For a visual sample $V_i$ with its original text $T_i$, let
$\mathcal C_i$ denote the candidate texts from other samples. A pretrained
encoder provides a similarity score $s(T_i,T_j)$. After excluding the
original source, duplicates, and near duplicates, we select the highest
ranked valid negative:
\begin{equation}
T_i^{-}
\in\operatorname*{arg\,max}_{T_j\in\mathcal C_i,\;j\neq i}
s(T_i,T_j).
\label{eq:app_hard_negative}
\end{equation}
Negative construction is performed independently within every partition.
The primary overall-reliability setting contains 50\% mismatched inputs.
We additionally evaluate mismatch ratios of 25\% and 75\% and compare hard,
random, and easy negatives. Mixed test sets are generated with seeds
$\{42,43,44,45,46\}$.

\subsection{Preprocessing and Generation Prompts}
\label{app:preprocessing_prompts}

For MINA, each source video is represented by 16 uniformly sampled RGB
frames resized to $336\times336$. The generation pipeline selects eight
evenly spaced frames from this precomputed 16-frame bank. The original
post is used without semantic rewriting. The fixed instruction asks the model to
act as a multimodal intention-understanding assistant, jointly analyze
the visual content and original post, and output exactly ten concise,
human-centric English intentions using ten prescribed prefixes. Each
completion is limited to 25 words, and introductions, explanations,
notes, bullets, and other extra text are prohibited. The original post is
inserted into the final prompt field. The required output prefixes are:
\begin{enumerate}
    \item \texttt{After posting this Tweet, the user wants to ...}
    \item \texttt{After viewing this Tweet, others will ...}
    \item \texttt{The user posts this Tweet because the user is ...}
    \item \texttt{The user posts this Tweet because the user intended to ...}
    \item \texttt{After posting this Tweet, the user feels ...}
    \item \texttt{After viewing this Tweet, others feel ...}
    \item \texttt{After viewing this Tweet, others want to ...}
    \item \texttt{After posting this Tweet, the user will ...}
    \item \texttt{Before posting this Tweet, the user needs to ...}
    \item \texttt{The user posts this Tweet because ...}
\end{enumerate}

For MIND, preprocessing first filters for
\texttt{agreement=yes} and retains the first 3,000 records in repository
order. At most three evenly selected product images are retained for each
side of a pair and resized with a maximum side length of 448 pixels. Each
product description is cleaned and limited to 1,800 characters.
The fixed instruction asks the model to infer the most plausible co-buy
intention from the two products' images, titles, descriptions, category,
and relation hint. The output must be one concise English sentence that
starts with ``The potential co-buy intention for these two products could
be that'' and explains why a user may buy the products together. Mentions
of uncertainty and any labels, bullets, introductions, or extra text are
prohibited. Prompts and preprocessing are fixed across uncertainty
estimators.

\subsection{Intention Generation and Semantic Targets}
\label{app:generation_targets}

We use Qwen3-VL-8B-Instruct~\cite{bai2025qwen3vl} as the generation
backbone. Deterministic
decoding produces the primary response, with maximum generation lengths
of 256 tokens for MINA and 96 tokens for MIND. DUE uses this single
autoregressive response and its hidden states.
Sampling-based baselines use the same input and instruction but generate
the additional responses required by their original uncertainty rules.

For response $Y_i$ and reference $Y_i^{*}$, BERTScore
~\cite{zhang2020bertscore} defines the continuous
semantic-error target as
\begin{equation}
q_i
=
1-\mathrm{BERTScore}(Y_i,Y_i^{*}).
\end{equation}
The binary semantic-error label is
\begin{equation}
e_i
=
\mathbb I[
\mathrm{BERTScore}(Y_i,Y_i^{*})<\tau],
\end{equation}
where the validation-selected threshold is $\tau=0.554039$. This threshold
is fixed before calibration and test evaluation. The continuous target
$q_i$ supervises uncertainty ranking, whereas $e_i$ is used for semantic
error detection and probability calibration.

\subsection{Overall-Reliability Protocol}
\label{app:overall_protocol}

The overall test event follows the sequential definition in the main
paper. A sample is unreliable when the multimodal input is mismatched, or
when semantic generation fails after receiving an aligned input. For each
mismatch ratio and mixture seed, all methods use the same source partition,
negative assignments, semantic labels, and evaluation examples. Aleatoric
and epistemic scores are calibrated separately on the held-out
calibration partition and are evaluated on the corresponding test mixture.

We report AUROC and AUPRC for discrimination, Spearman correlation
$\rho$ for continuous semantic-risk ranking, ECE and Brier score for
calibration, and AURC for selective prediction. ECE uses 10 equal-width
probability bins. Unless noted otherwise, aggregate results are
mean$\pm$standard deviation over five mixture seeds. Confidence intervals
use 1,000 bootstrap replicates with bootstrap seed 2027.

\subsection{Evaluation Metrics}
\label{app:evaluation_metrics}

AUROC measures the probability that a randomly selected failed sample is
ranked above a reliable sample~\cite{fawcett2006roc}, while AUPRC
summarizes the precision--recall trade-off and is informative under class
imbalance~\cite{davis2006relationship}. Spearman correlation
measures rank agreement between uncertainty and the continuous semantic
error target~\cite{spearman1904proof}. Probability quality is assessed by
the Brier score~\cite{brier1950verification} and 10-bin
ECE~\cite{guo2017calibration}, formally defined in
Section~\ref{app:calibration_protocol}.

For selective prediction, samples are sorted by increasing uncertainty.
At coverage $c$, selective risk $R(c)$ is the empirical failure rate
among the lowest-risk fraction $c$; AURC is the area under this
risk--coverage curve~\cite{geifman2017selective}. Thus, discrimination metrics evaluate error
separation, calibration metrics evaluate probabilistic accuracy, and
AURC evaluates whether uncertainty supports reliable rejection. We
report multiple complementary metrics because no single criterion fully
characterizes uncertainty quality.

\section{Implementation Details}
\label{app:implementation}

\subsection{Input-Side Aleatoric Estimator}
\label{app:data_side_details}

Following evidential deep learning~\cite{sensoy2018evidential}, the
aleatoric branch contains a shared-space vision--language encoder, an
evidential matching component, and a lightweight pairwise prediction head.
For each input pair, the branch uses the expected matching probability
$p_{\mathrm{match}}$, evidential uncertainty $u_{\mathrm{evi}}$, the
combined evidential alignment risk
$U_{\mathrm{ale}}^{\mathrm{evi}}$, pretrained embedding distance, and
video-level temporal statistics.

For video samples, frame--text similarities
$\{s_{it}\}_{t=1}^{F}$ can be summarized using:
\begin{equation}
\begin{aligned}
s_i^{\max}
&=
\max_t s_{it},
\\
s_i^{\mathrm{mean}}
&=
\frac{1}{F}\sum_{t=1}^{F}s_{it},
\\
s_i^{\mathrm{std}}
&=
\sqrt{
\frac{1}{F}\sum_{t=1}^{F}
(s_{it}-s_i^{\mathrm{mean}})^2
},
\\
d_i^{\mathrm{temp}}
&=
\frac{1}{F-1}
\sum_{t=2}^{F}
|s_{it}-s_{i,t-1}|.
\end{aligned}
\label{eq:app_temporal_features}
\end{equation}
The maximum and mean similarities measure key-frame and global support,
while the standard deviation and adjacent-frame difference characterize
cross-frame inconsistency and temporal fluctuation.

The two evidential outputs are constrained to be positive through
Softplus transformations. To avoid unstable gradients, all logarithms are
computed after numerical clipping, and the Dirichlet concentration
$S_{VT}$ is lower-bounded by a small positive constant. The pairwise head
combines evidential, representation-distance, and temporal cues into
$U_{\mathrm{ale}}^{\mathrm{raw}}$. Training follows the two stages
described in the main paper. First, the vision--language representations
and evidence head are optimized with symmetric contrastive, evidential
classification, and label-aware Dirichlet KL losses. The backbone is then
frozen and the pairwise risk head is trained with class-weighted binary
cross-entropy. Training uses 10 epochs, batch size 16, learning rate
$10^{-3}$, hidden dimension 128, dropout 0.05, and five negatives per
positive pair. All aleatoric baselines use identical source-level
partitions and pair manifests.

\subsection{Sentence-Level Epistemic Uncertainty}
\label{app:token_sequence_details}

For each generated token, hidden states from the final $K$ transformer
layers are cached during the same autoregressive response. Internal samples
are produced by drawing layer-mixture weights from a Dirichlet
distribution. The mixed representations are passed through the frozen
final normalization layer and language-modeling head, yielding $S$
predictive distributions for each token position. The final configuration
uses $K=5$, $S=10$, and a
symmetric Dirichlet distribution with concentration $\alpha=1.0$.

The proposed procedure does not repeat multimodal encoding or
autoregressive decoding. Its additional cost is limited to weighted
hidden-state combinations, language-modeling-head projections, and the
small sequence estimator. For fair efficiency reporting, the runtime must
include these additional operations.

Padding positions and prompt tokens are excluded from the uncertainty
sequence. Only generated content tokens are retained. When a response is
truncated, its final valid token is included, but padding introduced for
batching is masked in all sequence operations and losses.

\begin{table*}[t]
\centering
\scriptsize
\begin{tabular*}{\textwidth}{@{\extracolsep{\fill}}lll@{}}
\toprule
Component & Hyperparameter & Final value \\
\midrule
Generation & Base MLLM & Qwen3-VL-8B-Instruct \\
Generation & Responses used by DUE & $1$ \\
Generation & Maximum new tokens (MINA / MIND) & $256/96$ \\
Aleatoric & Epochs / batch size & $10/16$ \\
Aleatoric & Optimizer / learning rate & AdamW / $10^{-3}$ \\
Aleatoric & Weight decay & $10^{-4}$ \\
Aleatoric & Hidden dimension / dropout & $128/0.05$ \\
Aleatoric & Negatives per positive & $5$ \\
Epistemic & Mixed layers $K$ / internal mixtures $S$ & $5/10$ \\
Epistemic & Dirichlet concentration $\alpha$ & $1.0$ \\
Epistemic & Epochs / batch size & $100/32$ \\
Epistemic & Optimizer / learning rate & AdamW / $10^{-3}$ \\
Epistemic & Weight decay / gradient clipping & $10^{-5}/5.0$ \\
Epistemic & Hidden dimension / dropout & $96/0.1$ \\
Epistemic & Recurrent layers / direction & $1$ / bidirectional \\
Epistemic & Mask ratio / uncertainty weight $\eta$ & $0.3/1.0$ \\
Epistemic & Rank margin / minimum error gap & $0.05/0.02$ \\
Epistemic & $\lambda_{\mathrm{seq}}$ & $0.02$ \\
Epistemic & $\lambda_{\mathrm{cal}}$ & $0.05$ \\
Epistemic & $\lambda_{\mathrm{rank}}$ & $5.0$ \\
Epistemic & Selection metric & Spearman correlation \\
Training & Random seed & $42$ \\
Calibration & ECE bins & $10$ \\
Evaluation & Mixture seeds & $42, 43, 44, 45, 46$ \\
Evaluation & Bootstrap replicates / seed & $1{,}000/2027$ \\
\bottomrule
\end{tabular*}
\caption{Hyperparameters of the final configuration.}
\label{tab:app_hyperparameters}
\end{table*}

\subsection{Sequence Encoder and NIG Parameterization}
\label{app:nig_implementation}

Following NIG-based evidential regression~\cite{amini2020deep}, let
$\mathbf u_i=(u_{i1},\ldots,u_{iN_i})$ be the valid token uncertainty
sequence. A recurrent encoder receives token values, positional
information, and an attention mask. Its attention-pooled representation is
concatenated with global statistics such as the mean, maximum, standard
deviation, selected quantiles, and adjacent-token variation. A prediction
head outputs four unconstrained values
$(r_{\gamma},r_{\nu},r_{\alpha},r_{\beta})$, which are converted to valid
NIG parameters:
\begin{align}
\gamma &= r_{\gamma}, \nonumber\\
\nu &= \operatorname{Softplus}(r_{\nu})+\epsilon, \nonumber\\
\alpha &= \operatorname{Softplus}(r_{\alpha})+1+\epsilon, \nonumber\\
\beta &= \operatorname{Softplus}(r_{\beta})+\epsilon.
\label{eq:app_nig_constraints}
\end{align}
Here, $\epsilon>0$ prevents invalid or numerically singular parameter
values. The offset in $\alpha$ ensures $\alpha>1$, which is required for
the posterior variance of the latent mean to exist.

During masked uncertainty prediction, valid sequence positions are
randomly partitioned into observed and masked sets. The masked ratio is
sampled during training. At inference time, the complete valid sequence is
observed, and the NIG parameters are predicted without masking.

\subsection{Optimization and Reproducibility}
\label{app:optimization}

Table~\ref{tab:app_hyperparameters} summarizes the final configuration.
Entropy values are normalized using training-set statistics. Model
selection is performed on the validation partition using Spearman
correlation between the raw epistemic score and semantic error.

Training uses seed 42, which controls data shuffling, negative sampling,
layer-mixture sampling, masking, and parameter initialization. Mixed-set
seeds alter only the construction of evaluation mixtures. The same
mixtures are reused by all compared fusion methods. Candidate epistemic
configurations varied encoder width, dropout, loss weights, and the
validation selection metric; the final configuration was selected by
validation Spearman correlation. The released experiment configuration
records all evaluated candidates.

\subsection{Baseline Implementation}
\label{app:baseline_implementation}

The aleatoric comparison includes PCME++~\cite{chun2024improved},
UATVR~\cite{fang2023uatvr}, PAU~\cite{li2023prototype}, and
FUME~\cite{duan2025fuzzy}. We use
their public implementations and native probabilistic or uncertainty
scores, followed by the same held-out monotonic calibration used for our
aleatoric score. PCME++ uses its closed-form probabilistic similarity,
UATVR uses its probabilistic text--video matching score, PAU uses its
prototype-based uncertainty, and FUME uses fuzzy cross-modal uncertainty
to modulate pair similarity.

The epistemic comparison includes token-entropy statistics
~\cite{malinin2021uncertainty},
SelfCheckGPT~\cite{manakul2023selfcheckgpt}, Semantic
Entropy~\cite{kuhn2023semantic}, P(True)~\cite{kadavath2022language},
and VL-Uncertainty~\cite{zhang2024vluncertainty}. Entropy
statistics use the primary response. P(True) requests an additional
confidence judgment, SelfCheckGPT and Semantic Entropy use five additional
sampled responses, and VL-Uncertainty uses four multimodal perturbation
responses. All raw baseline scores are oriented so that a larger value
indicates greater semantic-error risk and are calibrated only on the
held-out calibration partition.

\section{Additional Mathematical Details}
\label{app:mathematical_details}

\subsection{NIG Posterior Predictive Distribution}
\label{app:nig_predictive}

Following deep evidential regression~\cite{amini2020deep}, the
hierarchical model for a masked token uncertainty $u$ is
\begin{equation}
\begin{aligned}
\sigma^2
&\sim
\operatorname{InvGamma}(\alpha,\beta),
\\
\mu\mid\sigma^2
&\sim
\mathcal N\left(\gamma,\frac{\sigma^2}{\nu}\right),
\\
u\mid\mu,\sigma^2
&\sim
\mathcal N(\mu,\sigma^2).
\end{aligned}
\end{equation}
Marginalizing $\mu$ and $\sigma^2$ yields a Student-$t$ predictive
distribution:
\begin{equation}
u
\sim
\operatorname{StudentT}
\left(
2\alpha,\,
\gamma,\,
\sqrt{
\frac{\beta(1+\nu)}
{\alpha\nu}
}
\right).
\label{eq:app_student_t}
\end{equation}
Therefore, the masked-sequence objective can be implemented as the sum of
Student-$t$ negative log-likelihoods over valid masked positions. This
heavy-tailed predictive distribution is more tolerant of localized token
uncertainty peaks than a fixed-variance Gaussian likelihood.

The posterior mean of the latent sentence uncertainty is $\gamma$. Its
variance is
\begin{equation}
\operatorname{Var}(\mu\mid\mathbf u)
=
\mathbb E_{\sigma^2}
\left[
\frac{\sigma^2}{\nu}
\right]
=
\frac{\beta}{\nu(\alpha-1)}.
\end{equation}
Consequently, the conservative score
$\gamma+\eta\sqrt{\beta/[\nu(\alpha-1)]}$ increases either when the
estimated sentence uncertainty is high or when the estimate itself is
insufficiently supported.

\subsection{Why Evidential Uncertainty Alone Is Insufficient}
\label{app:evidential_explanation}

For a two-class Dirichlet distribution, $u_{\mathrm{evi}}=2/S$ depends on
the total evidence $S$ but not on which relation receives that evidence.
Thus, a confidently matched pair and a confidently mismatched pair may
both have low evidential uncertainty. The expected matching probability
$p_{\mathrm{match}}$ supplies the missing direction of evidence. The
combined score
\begin{equation}
U_{\mathrm{ale}}^{\mathrm{evi}}
=
(1-p_{\mathrm{match}})
+
p_{\mathrm{match}}u_{\mathrm{evi}}
\end{equation}
has two interpretable components: explicit mismatch risk and residual
ignorance under an apparently matched relation.

Its limiting behavior is also intuitive. When
$p_{\mathrm{match}}\rightarrow 0$, the score approaches one regardless of
the evidence amount. When $p_{\mathrm{match}}\rightarrow 1$, the score
reduces to $u_{\mathrm{evi}}$, so a pair is regarded as reliable only when
the evidence both supports matching and is sufficiently concentrated.

\subsection{Sequential Failure Decomposition}
\label{app:fusion_derivation}

Let $A=\mathcal E_{\mathrm{align}}$ be the input-failure event and let
$M=\mathcal E_{\mathrm{sem}}$ be semantic failure. Assigning each failed
sample to its earliest failure stage gives
\begin{equation}
\mathcal E_{\mathrm{total}}
=
A\cup(\neg A\cap M).
\end{equation}
Because $A$ and $\neg A\cap M$ are disjoint,
\begin{align}
P(\mathcal E_{\mathrm{total}})
&=
P(A)+P(\neg A\cap M)\nonumber\\
&=
P(A)+P(\neg A)P(M\mid\neg A).
\label{eq:app_total_probability}
\end{align}
Substituting the calibrated estimates
$\widetilde U_{\mathrm{ale}}\approx P(A)$ and
$\widetilde U_{\mathrm{epi}}\approx P(M\mid\neg A)$ gives
\begin{equation}
U_{\mathrm{total}}
=
\widetilde U_{\mathrm{ale}}
+
(1-\widetilde U_{\mathrm{ale}})
\widetilde U_{\mathrm{epi}}.
\end{equation}
This is algebraically equivalent to
\begin{equation}
U_{\mathrm{total}}
=
1-
(1-\widetilde U_{\mathrm{ale}})
(1-\widetilde U_{\mathrm{epi}}).
\end{equation}
The decomposition does not assume unconditional independence between input
mismatch and output error. Instead, the epistemic term is explicitly
conditional on survival of the input stage.

The fusion also satisfies
\begin{equation}
U_{\mathrm{total}}
\geq
\max(
\widetilde U_{\mathrm{ale}},
\widetilde U_{\mathrm{epi}}
),
\end{equation}
for calibrated probabilities in $[0,1]$. Hence, a high failure
probability from either stage cannot be averaged away by a low probability
from the other stage.

\section{Calibration Protocol}
\label{app:calibration_protocol}

The two calibrators are fitted independently because the branches estimate
different events. The aleatoric calibrator uses all calibration samples
and their input-match labels. The epistemic calibrator uses only aligned
calibration samples and their semantic-error labels. This restriction is
necessary for the calibrated epistemic output to approximate
$P(\mathcal E_{\mathrm{sem}}\mid\neg\mathcal E_{\mathrm{align}})$.

We use isotonic regression~\cite{zadrozny2002transforming} for both
branches. Before fitting, the score
direction is selected on calibration data so that larger transformed values
consistently indicate higher failure risk. The fitted mappings are then
applied unchanged to the test set:
\begin{equation}
\widetilde U_{\mathrm{ale}}
=f_{\mathrm{ale}}(U_{\mathrm{ale}}^{\mathrm{raw}}),
\qquad
\widetilde U_{\mathrm{epi}}
=f_{\mathrm{epi}}(U_{\mathrm{epi}}^{\mathrm{raw}}).
\end{equation}
Aleatoric calibration is performed at the unique-pair level, preventing
repeated intention candidates from over-weighting the same visual--textual
relation. Epistemic calibration is performed at the intention-item level
on aligned inputs.

Calibration quality is evaluated using ECE and Brier score. For $N$
samples with predicted failure probabilities $\widehat p_i$ and binary
labels $y_i$, the Brier score is
\begin{equation}
\operatorname{Brier}
=
\frac{1}{N}
\sum_{i=1}^{N}
(\widehat p_i-y_i)^2.
\end{equation}
For $J$ probability bins $\{B_j\}_{j=1}^{J}$, ECE is
\begin{equation}
\operatorname{ECE}
=
\sum_{j=1}^{J}
\frac{|B_j|}{N}
\left|
\operatorname{acc}(B_j)
-
\operatorname{conf}(B_j)
\right|.
\end{equation}
We use $J=10$ equal-width bins throughout.

\begin{table*}[t]
\centering
\scriptsize
\setlength{\tabcolsep}{2.0pt}
\renewcommand{\arraystretch}{0.97}
\begin{tabular*}{\textwidth}{
@{\extracolsep{\fill}}lllccccc@{}}
\toprule
Dataset & Aleatoric Method & Epistemic Method & AUROC$\uparrow$
& AUPRC$\uparrow$ & ECE$\downarrow$ & Brier$\downarrow$
& AURC$\downarrow$ \\
\midrule
\multirow{10}{*}{MINA}
& PCME++ & P(True) & $0.678\pm0.016$ & $0.883\pm0.009$
& $0.058\pm0.016$ & $0.176\pm0.008$ & $0.670\pm0.009$ \\
& PCME++ & SelfCheckGPT & $0.680\pm0.024$ & $0.883\pm0.011$
& $\underline{0.052\pm0.019}$ & $0.175\pm0.009$ & $0.668\pm0.014$ \\
& UATVR & Semantic Entropy & $0.710\pm0.025$ & $0.884\pm0.011$
& $0.057\pm0.009$ & $0.171\pm0.005$ & $0.644\pm0.024$ \\
& UATVR & SelfCheckGPT & $0.711\pm0.024$ & $0.884\pm0.011$
& $0.052\pm0.011$ & $0.170\pm0.003$ & $0.644\pm0.024$ \\
& PAU & P(True) & $0.756\pm0.026$ & $0.912\pm0.013$
& $0.063\pm0.022$ & $0.163\pm0.006$ & $0.622\pm0.024$ \\
& PAU & VL-Uncertainty & $0.757\pm0.026$ & $0.912\pm0.010$
& $0.054\pm0.019$ & $0.161\pm0.007$ & $\underline{0.620\pm0.025}$ \\
& PAU & SelfCheckGPT & $0.759\pm0.028$ & $0.912\pm0.010$
& $0.057\pm0.019$ & $0.161\pm0.007$ & $0.621\pm0.028$ \\
& FUME & Semantic Entropy & $0.756\pm0.020$ & $0.915\pm0.008$
& $0.046\pm0.016$ & $0.161\pm0.009$ & $0.626\pm0.017$ \\
& FUME & SelfCheckGPT & $\underline{0.759\pm0.017}$
& $\underline{0.916\pm0.009}$ & $\mathbf{0.043\pm0.012}$
& $\underline{0.160\pm0.008}$ & $0.621\pm0.021$ \\
& \textbf{$U_{\mathrm{ale}}$ (Ours)}
& \textbf{$U_{\mathrm{epi}}$ (Ours)}
& $\mathbf{0.806\pm0.013}$ & $\mathbf{0.935\pm0.005}$
& $0.059\pm0.011$ & $\mathbf{0.150\pm0.007}$
& $\mathbf{0.596\pm0.018}$ \\
\midrule
\multirow{10}{*}{MIND}
& PCME++ & Semantic Entropy & $0.621\pm0.027$ & $0.860\pm0.018$
& $0.067\pm0.021$ & $0.180\pm0.016$ & $0.715\pm0.032$ \\
& PCME++ & P(True) & $0.624\pm0.020$ & $0.860\pm0.018$
& $0.084\pm0.028$ & $0.181\pm0.015$ & $0.708\pm0.019$ \\
& FUME & P(True) & $0.655\pm0.022$ & $0.869\pm0.018$
& $0.091\pm0.033$ & $0.179\pm0.016$ & $0.690\pm0.027$ \\
& FUME & Semantic Entropy & $0.656\pm0.029$ & $0.870\pm0.014$
& $0.079\pm0.023$ & $0.178\pm0.016$ & $0.691\pm0.029$ \\
& UATVR & P(True) & $0.685\pm0.027$ & $0.890\pm0.013$
& $0.061\pm0.011$ & $0.168\pm0.008$ & $0.679\pm0.037$ \\
& UATVR & Semantic Entropy & $0.686\pm0.021$ & $0.889\pm0.010$
& $\mathbf{0.058\pm0.015}$ & $0.169\pm0.009$ & $0.680\pm0.033$ \\
& PAU & SelfCheckGPT & $0.707\pm0.015$ & $0.897\pm0.007$
& $0.064\pm0.012$ & $0.165\pm0.010$ & $0.663\pm0.024$ \\
& PAU & Semantic Entropy & $0.712\pm0.014$ & $0.898\pm0.008$
& $\underline{0.059\pm0.010}$ & $0.166\pm0.012$ & $0.662\pm0.028$ \\
& PAU & P(True) & $\underline{0.721\pm0.020}$
& $\underline{0.900\pm0.013}$ & $0.066\pm0.020$
& $\underline{0.164\pm0.010}$ & $\underline{0.657\pm0.028}$ \\
& \textbf{$U_{\mathrm{ale}}$ (Ours)}
& \textbf{$U_{\mathrm{epi}}$ (Ours)}
& $\mathbf{0.755\pm0.027}$ & $\mathbf{0.923\pm0.003}$
& $0.061\pm0.014$ & $\mathbf{0.158\pm0.007}$
& $\mathbf{0.647\pm0.045}$ \\
\bottomrule
\end{tabular*}
\caption{Representative overall-reliability comparisons under a 50\%
input-mismatch setting. Results are mean $\pm$ standard deviation over
five mixture seeds.}
\label{tab:app_total_combinations}
\end{table*}

\section{Complete Quantitative Results}
\label{app:complete_results}

\subsection{Aleatoric and Epistemic Results}

Tables~\ref{tab:app_input_results} and~\ref{tab:app_model_results} report
the complete component-level comparisons used in the paper. The
aleatoric results evaluate cross-modal mismatch detection, while the
epistemic results evaluate semantic-error detection and continuous error
ranking. The proposed components provide the strongest discrimination and
selective-prediction performance on both datasets.

\begin{table}[t]
\centering
\scriptsize
\setlength{\tabcolsep}{0.8pt}
\renewcommand{\arraystretch}{0.92}
\begin{tabular*}{\columnwidth}{
@{\extracolsep{\fill}}llccccc@{}}
\toprule
Dataset & Method & AUROC$\uparrow$ & AUPRC$\uparrow$
& ECE$\downarrow$ & Brier$\downarrow$ & AURC$\downarrow$ \\
\midrule
\multirow{5}{*}{MINA}
& PCME++ & $0.8028$ & $0.8187$ & $\mathbf{0.0438}$ & $0.1847$ & $0.3009$ \\
& UATVR & $0.8337$ & $0.8270$ & $\underline{0.0566}$ & $0.1679$ & $0.2775$ \\
& PAU & $\underline{0.8922}$ & $0.8718$ & $0.0945$
& $\underline{0.1387}$ & $\underline{0.2403}$ \\
& FUME & $0.8780$ & $\underline{0.8895}$ & $0.0699$ & $0.1517$ & $0.2497$ \\
& $U_{\mathrm{ale}}$ (Ours) & $\mathbf{0.9235}$ & $\mathbf{0.9340}$
& $0.0868$ & $\mathbf{0.1223}$ & $\mathbf{0.2298}$ \\
\midrule
\multirow{5}{*}{MIND}
& PCME++ & $0.7283$ & $0.7511$ & $0.0419$ & $0.2141$ & $0.3681$ \\
& UATVR & $0.8053$ & $0.8105$ & $0.0453$ & $0.1830$ & $0.3048$ \\
& PAU & $\underline{0.8351}$ & $\underline{0.8433}$ & $\underline{0.0382}$
& $\underline{0.1674}$ & $\underline{0.2816}$ \\
& FUME & $0.8083$ & $0.7995$ & $\mathbf{0.0369}$ & $0.1787$ & $0.2920$ \\
& $U_{\mathrm{ale}}$ (Ours) & $\mathbf{0.8942}$ & $\mathbf{0.9020}$
& $\mathbf{0.0369}$ & $\mathbf{0.1365}$ & $\mathbf{0.2382}$ \\
\bottomrule
\end{tabular*}
\caption{Complete input-side cross-modal mismatch detection results.}
\label{tab:app_input_results}
\end{table}

\begin{table}[t]
\centering
\scriptsize
\setlength{\tabcolsep}{0.4pt}
\renewcommand{\arraystretch}{0.92}
\begin{tabular*}{\columnwidth}{
@{\extracolsep{\fill}}llcccccc@{}}
\toprule
Dataset & Method & AUROC$\uparrow$ & AUPRC$\uparrow$ & $\rho$$\uparrow$
& ECE$\downarrow$ & AURC$\downarrow$ & Time$\downarrow$ \\
\midrule
\multirow{7}{*}{MINA}
& Entropy Mean & $\underline{0.578}$ & $\underline{0.598}$
& $\underline{0.163}$ & $0.036$ & $\underline{0.476}$ & $\mathbf{1.0\times}$ \\
& Entropy Std & $0.561$ & $0.596$ & $0.120$ & $\mathbf{0.012}$
& $0.488$ & $\mathbf{1.0\times}$ \\
& SelfCheckGPT & $0.509$ & $0.532$ & $0.037$ & $0.037$ & $0.512$ & $6.0\times$ \\
& Semantic Entropy & $0.496$ & $0.528$ & $0.012$ & $0.041$ & $0.527$ & $6.0\times$ \\
& P(True) & $0.506$ & $0.538$ & $-0.079$ & $0.019$ & $0.516$
& $\underline{1.9\times}$ \\
& VL-Uncertainty & $0.507$ & $0.528$ & $0.026$ & $\underline{0.018}$
& $0.512$ & $5.0\times$ \\
& $U_{\mathrm{epi}}$ (Ours) & $\mathbf{0.631}$ & $\mathbf{0.648}$
& $\mathbf{0.265}$ & $0.049$ & $\mathbf{0.440}$ & $\mathbf{1.0\times}$ \\
\midrule
\multirow{7}{*}{MIND}
& Entropy Mean & $0.514$ & $0.507$ & $0.027$ & $0.042$ & $0.493$ & $\mathbf{1.0\times}$ \\
& Entropy Std & $0.496$ & $0.494$ & $0.010$ & $0.050$ & $0.500$ & $\mathbf{1.0\times}$ \\
& SelfCheckGPT & $\underline{0.538}$ & $\underline{0.541}$
& $\underline{0.061}$ & $\mathbf{0.010}$ & $0.487$ & $6.0\times$ \\
& Semantic Entropy & $0.525$ & $\underline{0.541}$ & $0.039$ & $0.102$
& $0.487$ & $6.0\times$ \\
& P(True) & $0.534$ & $0.527$ & $0.048$ & $0.040$ & $0.485$
& $\underline{1.9\times}$ \\
& VL-Uncertainty & $0.531$ & $0.539$ & $0.040$ & $\underline{0.013}$
& $\underline{0.481}$ & $5.0\times$ \\
& $U_{\mathrm{epi}}$ (Ours) & $\mathbf{0.548}$ & $\mathbf{0.551}$
& $\mathbf{0.099}$ & $0.039$ & $\mathbf{0.469}$ & $\mathbf{1.0\times}$ \\
\bottomrule
\end{tabular*}
\caption{Complete sentence-level epistemic semantic-error detection results.
$\rho$ denotes Spearman correlation.}
\label{tab:app_model_results}
\end{table}

\subsection{Representative Aleatoric--Epistemic Combinations}

Table~\ref{tab:app_total_combinations} reports an expanded comparison set
for the primary 50\% mismatch setting. Every row uses
separate event-probability calibration followed by Sequential Noisy-OR.
Hybrid rows containing only one proposed component are excluded, so the
comparison is between DUE and fully third-party aleatoric--epistemic
combinations.

\section{Additional Ablation Studies}
\label{app:ablations}

\subsection{Epistemic Component Ablation}

Table~\ref{tab:app_model_ablation} isolates the effects of attention,
global sequence statistics, sequence supervision, and ranking
supervision. Removing ranking supervision causes the largest degradation,
while the remaining components provide complementary improvements in
discrimination, ranking, and selective prediction.

\begin{table}[h]
\centering
\scriptsize
\setlength{\tabcolsep}{2.0pt}
\renewcommand{\arraystretch}{0.88}
\setlength{\aboverulesep}{0.3ex}
\setlength{\belowrulesep}{0.3ex}
\resizebox{\columnwidth}{!}{%
\begin{tabular}{lcccccc}
\toprule
Variant
& AUROC$\uparrow$
& AUPRC$\uparrow$
& Spearman$\uparrow$
& ECE$\downarrow$
& Brier$\downarrow$
& AURC$\downarrow$ \\
\midrule
Full
& $\mathbf{0.676}$ & $\mathbf{0.670}$ & $\mathbf{0.362}$
& $0.065$ & $\mathbf{0.229}$ & $\mathbf{0.384}$ \\
w/o Attention
& $0.669$ & $0.657$ & $\underline{0.353}$
& $\mathbf{0.049}$ & $0.231$ & $0.390$ \\
w/o Statistical Features
& $\underline{0.670}$ & $\underline{0.665}$ & $0.347$
& $\underline{0.049}$ & $\underline{0.230}$ & $\underline{0.389}$ \\
w/o Sequence Loss
& $0.658$ & $0.648$ & $0.324$
& $0.052$ & $0.236$ & $0.401$ \\
w/o Ranking Loss
& $0.638$ & $0.621$ & $0.284$
& $0.056$ & $0.239$ & $0.410$ \\
\bottomrule
\end{tabular}%
}
\caption{Epistemic component ablation on MINA using a fixed training
seed.}
\label{tab:app_model_ablation}
\end{table}

\begin{table*}[t]
\centering
\scriptsize
\setlength{\tabcolsep}{2.2pt}
\renewcommand{\arraystretch}{0.98}
\begin{tabular*}{\textwidth}{
@{\extracolsep{\fill}}llccccc@{}}
\toprule
Analysis & Method & AUROC$\uparrow$ & AUPRC$\uparrow$
& ECE$\downarrow$ & Brier$\downarrow$ & AURC$\downarrow$ \\
\midrule
\multirow{3}{*}{Decomposition}
& Epistemic-only $U_{\mathrm{epi}}$
& $0.588\pm0.009$
& $0.844\pm0.007$
& $\underline{0.030\pm0.008}$
& $0.184\pm0.004$
& $0.708\pm0.012$ \\

& Aleatoric-only $U_{\mathrm{ale}}$
& $0.790\pm0.008$
& $0.937\pm0.004$
& $0.084\pm0.014$
& $0.159\pm0.008$
& $0.607\pm0.011$ \\

& Full $U_{\mathrm{total}}$
& $\mathbf{0.806\pm0.011}$
& $0.935\pm0.005$
& $0.059\pm0.010$
& $\underline{0.150\pm0.006}$
& $\mathbf{0.596\pm0.016}$ \\
\midrule

\multirow{9}{*}{Fusion}
& Raw Sum + Late Isotonic
& $0.791\pm0.011$
& $\underline{0.938\pm0.004}$
& $0.076\pm0.010$
& $0.157\pm0.008$
& $0.605\pm0.014$ \\

& Raw Weighted Sum + Late Isotonic
& $0.801\pm0.017$
& $\mathbf{0.940\pm0.005}$
& $0.060\pm0.018$
& $0.152\pm0.010$
& $\mathbf{0.596\pm0.017}$ \\

& Raw Maximum + Late Isotonic
& $0.786\pm0.013$
& $0.937\pm0.003$
& $\mathbf{0.029\pm0.007}$
& $0.151\pm0.003$
& $0.609\pm0.019$ \\

& Calibrated Average
& $0.799\pm0.013$
& $0.930\pm0.006$
& $0.264\pm0.009$
& $0.221\pm0.009$
& $0.600\pm0.017$ \\

& Separate Calibration + Weighted Sum
& $0.801\pm0.012$
& $0.933\pm0.006$
& $0.260\pm0.011$
& $0.227\pm0.013$
& $0.600\pm0.016$ \\

& Learned MLP Fusion
& $0.800\pm0.014$
& $0.932\pm0.006$
& $0.086\pm0.010$
& $0.158\pm0.007$
& $0.599\pm0.016$ \\

& Noisy-OR w/o Conditional Calibration
& $\underline{0.804\pm0.011}$
& $0.935\pm0.004$
& $0.120\pm0.010$
& $0.169\pm0.006$
& $\underline{0.597\pm0.016}$ \\

& Shared Calibration + Noisy-OR
& $0.803\pm0.010$
& $0.935\pm0.004$
& $0.049\pm0.007$
& $\mathbf{0.149\pm0.004}$
& $0.598\pm0.016$ \\

& Sequential Noisy-OR (Ours)
& $\mathbf{0.806\pm0.011}$
& $0.935\pm0.005$
& $0.059\pm0.010$
& $\underline{0.150\pm0.006}$
& $\mathbf{0.596\pm0.016}$ \\
\bottomrule
\end{tabular*}
\caption{Aleatoric--epistemic decomposition and extended fusion analysis
on MINA under the 50\% mismatch setting. Best and second-best results are
shown in bold and underlined, respectively.}
\label{tab:app_fusion_ablation}
\end{table*}

\subsection{Aleatoric--Epistemic Decomposition and Extended Fusion Analysis}

Table~\ref{tab:app_fusion_ablation} separates the contribution of the two
risk sources and compares a broader set of fusion rules. Late-calibration
baselines first combine raw scores and then fit isotonic regression.
Separately calibrated methods first map both components to probabilities.
Although calibrated maximum is strong on some discrimination metrics, it
does not follow the sequential event definition. Sequential Noisy-OR
provides the intended probability semantics while maintaining a favorable
balance across discrimination, calibration, and selective prediction.

\subsection{Sensitivity to Mismatch Prevalence}

Table~\ref{tab:app_mismatch_ratio} varies the mismatch prevalence while
holding the component models and calibration protocol fixed.
$U_{\mathrm{total}}$ consistently improves over either component. The
margin is largest when both failure sources occur frequently enough to
contribute complementary information.

\begin{table}[h]
\centering
\scriptsize
\setlength{\tabcolsep}{0.8pt}
\renewcommand{\arraystretch}{0.96}
\begin{tabular*}{\columnwidth}{
@{\extracolsep{\fill}}llccc@{}}
\toprule
Ratio & Method & AUROC$\uparrow$ & Brier$\downarrow$ & AURC$\downarrow$ \\
\midrule
\multirow{3}{*}{$25\%$}
& Epistemic $U_{\mathrm{epi}}$ & $0.597\pm0.010$ & $0.229\pm0.001$
& $0.564\pm0.017$ \\
& Aleatoric $U_{\mathrm{ale}}$ & $\underline{0.685\pm0.013}$
& $\underline{0.208\pm0.003}$ & $\underline{0.524\pm0.014}$ \\
& Full $U_{\mathrm{total}}$ & $\mathbf{0.730\pm0.009}$
& $\mathbf{0.198\pm0.004}$ & $\mathbf{0.487\pm0.009}$ \\
\midrule
\multirow{3}{*}{$50\%$}
& Epistemic $U_{\mathrm{epi}}$ & $0.588\pm0.010$ & $0.184\pm0.005$
& $0.708\pm0.013$ \\
& Aleatoric $U_{\mathrm{ale}}$ & $\underline{0.790\pm0.009}$
& $\underline{0.159\pm0.008}$ & $\underline{0.607\pm0.013}$ \\
& Full $U_{\mathrm{total}}$ & $\mathbf{0.806\pm0.013}$
& $\mathbf{0.150\pm0.007}$ & $\mathbf{0.596\pm0.018}$ \\
\midrule
\multirow{3}{*}{$75\%$}
& Epistemic $U_{\mathrm{epi}}$ & $0.564\pm0.037$ & $0.107\pm0.003$
& $0.856\pm0.018$ \\
& Aleatoric $U_{\mathrm{ale}}$ & $\underline{0.864\pm0.009}$
& $\underline{0.090\pm0.003}$ & $\underline{0.759\pm0.013}$ \\
& Full $U_{\mathrm{total}}$ & $\mathbf{0.871\pm0.008}$
& $\mathbf{0.087\pm0.002}$ & $\mathbf{0.756\pm0.010}$ \\
\bottomrule
\end{tabular*}
\caption{Sensitivity to input-mismatch prevalence on MINA over five
mixture seeds.}
\label{tab:app_mismatch_ratio}
\end{table}

\subsection{Negative-Construction Robustness}

Hard negatives are substantially more challenging than random or easy
negatives. Table~\ref{tab:app_negative_robustness} shows that the complete
framework remains effective under the hard setting used for the primary
evaluation and improves further when mismatch cues are less ambiguous.

\begin{table}[t]
\centering
\scriptsize
\setlength{\tabcolsep}{2.3pt}
\begin{tabular*}{\columnwidth}{
@{\extracolsep{\fill}}lccccc@{}}
\toprule
Negative & AUROC$\uparrow$ & AUPRC$\uparrow$ & ECE$\downarrow$
& Brier$\downarrow$ & AURC$\downarrow$ \\
\midrule
Hard & $0.806$ & $0.935$ & $0.059$ & $0.150$ & $0.596$ \\
Random & $0.862$ & $0.957$ & $0.025$ & $0.125$ & $0.562$ \\
Easy & $0.853$ & $0.954$ & $0.034$ & $0.130$ & $0.566$ \\
\bottomrule
\end{tabular*}
\caption{Robustness to mismatch construction on MINA.}
\label{tab:app_negative_robustness}
\end{table}

\begin{table}[t]
\centering
\scriptsize
\setlength{\tabcolsep}{2.2pt}
\begin{tabular*}{\columnwidth}{
@{\extracolsep{\fill}}lccccc@{}}
\toprule
$\tau$ & AUROC$\uparrow$ & AUPRC$\uparrow$ & ECE$\downarrow$
& Brier$\downarrow$ & AURC$\downarrow$ \\
\midrule
$0.500000$ & $0.842$ & $0.916$ & $0.105$ & $0.164$ & $0.408$ \\
$0.554039$ & $0.806$ & $0.935$ & $0.059$ & $0.150$ & $0.596$ \\
$0.600000$ & $0.795$ & $0.959$ & $0.027$ & $0.112$ & $0.740$ \\
\bottomrule
\end{tabular*}
\caption{Sensitivity of $U_{\mathrm{total}}$ to the BERTScore threshold
on MINA over five mixture seeds.}
\label{tab:app_threshold}
\end{table}

\begin{table}[t]
\centering
\scriptsize
\setlength{\tabcolsep}{1.4pt}
\resizebox{\columnwidth}{!}{%
\begin{tabular}{@{}lcccc@{}}
\toprule
Method & AUROC & AUPRC & Brier & AURC \\
\midrule
Epistemic $U_{\mathrm{epi}}$
& $[0.559, 0.689]$ & $[0.739, 0.921]$ & $[0.176, 0.226]$ & $[0.530, 0.831]$ \\
Aleatoric $U_{\mathrm{ale}}$
& $[0.542, 0.833]$ & $[0.779, 0.965]$ & $[0.127, 0.254]$ & $[0.516, 0.825]$ \\
Full $U_{\mathrm{total}}$
& $[0.625, 0.883]$ & $[0.816, 0.976]$ & $[0.122, 0.262]$ & $[0.459, 0.786]$ \\
\bottomrule
\end{tabular}}
\caption{Bootstrap 95\% confidence intervals on MINA.}
\label{tab:app_bootstrap}
\end{table}

\subsection{Semantic-Threshold Sensitivity}

The semantic threshold changes both the definition and prevalence of the
overall failure event. As shown in Table~\ref{tab:app_threshold}, DUE
maintains useful discrimination and probability quality across all three
validation-relevant thresholds. The primary threshold is 0.554.

\subsection{Bootstrap Confidence Intervals}

Table~\ref{tab:app_bootstrap} reports 95\% intervals from 1,000 bootstrap
replicates for the primary 50\% mismatch setting. The intervals complement
the across-mixture standard deviations reported in the main tables.

\section{Additional Efficiency Analysis}
\label{app:efficiency}

The runtime comparison separates multimodal encoding, autoregressive
decoding, and post-generation uncertainty computation.
Let $T_{\mathrm{enc}}$, $T_{\mathrm{dec}}$, and $T_{\mathrm{unc}}$ denote
these three costs. For the proposed method,
\begin{equation}
T_{\mathrm{ours}}
=
T_{\mathrm{enc}}
+
T_{\mathrm{dec}}
+
T_{\mathrm{unc}},
\end{equation}
where $T_{\mathrm{unc}}$ contains only non-autoregressive operations.
For a sampling-based method using $R$ additional responses, the cost is
approximately
\begin{equation}
T_{\mathrm{sample}}
\approx
(R+1)
(T_{\mathrm{enc}}+T_{\mathrm{dec}})
+
T_{\mathrm{compare}},
\end{equation}
unless visual features are explicitly cached by that baseline.
Table~\ref{tab:app_efficiency} reports the resulting relative costs.

\begin{table}[h]
\centering
\scriptsize
\setlength{\tabcolsep}{4.0pt}
\begin{tabular*}{\columnwidth}{
@{\extracolsep{\fill}}lccc@{}}
\toprule
Method & Responses & Extra decoding & Rel. time$\downarrow$ \\
\midrule
Entropy statistics & $1$ & No & $1.0\times$ \\
P(True) & $2$ & Yes & $1.9\times$ \\
VL-Uncertainty & $5$ & Yes & $5.0\times$ \\
SelfCheckGPT & $6$ & Yes & $6.0\times$ \\
Semantic Entropy & $6$ & Yes & $6.0\times$ \\
$U_{\mathrm{epi}}$ (Ours) & $1$ & No & $1.0\times$ \\
\bottomrule
\end{tabular*}
\caption{Inference cost relative to one deterministic generation.
``Responses'' includes the primary intention response.}
\label{tab:app_efficiency}
\end{table}

Across 1,244 timed samples, one deterministic generation required
93.01 seconds on average in the measurement environment. Five sampled
responses required 465.47 seconds on average. These measurements confirm
that repeated decoding dominates the cost, whereas DUE reuses cached
states and adds only non-autoregressive projections and sequence inference.

\begin{figure}[h!]
    \centering
    \includegraphics[width=\columnwidth]
    {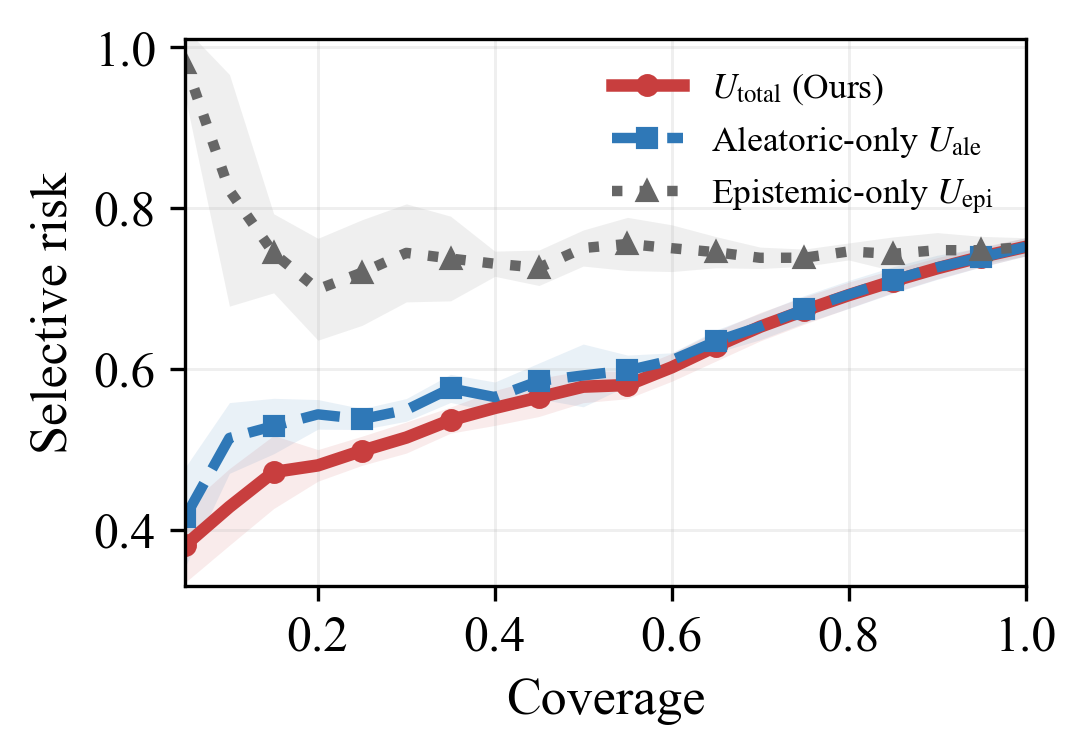}
    \caption{Risk--coverage curves on MINA under the 50\% input-mismatch
    setting. Shaded regions denote standard deviations over five mixture
    seeds.}
    \label{fig:app_risk_coverage}
\end{figure}

\section{Selective Prediction Analysis}\label{app:selective_prediction}
Risk--coverage evaluation sorts test samples by predicted uncertainty and
retains the least uncertain fraction at each coverage level. The selective
risk is the empirical failure rate among retained samples. As shown in
Figure~\ref{fig:app_risk_coverage}, $U_{\mathrm{total}}$ yields lower risk
than either component alone over low and medium coverage, where selective
prediction is most useful. The shaded regions show variability across the
five mixture seeds.

\fi

\end{document}